\documentclass[10pt,twocolumn,letterpaper]{article}

\usepackage{cvpr}              
\usepackage[accsupp]{axessibility}

\usepackage{float}
\usepackage{graphicx}
\usepackage{booktabs}
\usepackage{array}
\usepackage{tabularx}
\usepackage{caption}
\usepackage{multirow}
\usepackage{placeins}
\usepackage{afterpage}
\usepackage{tikz}
\usepackage{pgf-pie} 

\definecolor{cvprblue}{rgb}{0.21,0.49,0.74}
\usepackage[pagebackref,breaklinks,colorlinks,allcolors=cvprblue]{hyperref}

\def\paperID{} 
\def\confName{CVPR}
\def\confYear{2026}

\title{Camera Control for Text-to-Image Generation via Learning Viewpoint Tokens}

\author{
Xinxuan Lu \quad Charless Fowlkes \quad Alexander C. Berg \\
University of California, Irvine \\
{\tt\small \{xinxul1, fowlkes, bergac\}@uci.edu}
}

\begin{document}
\maketitle
\begin{abstract}
Current text-to-image models struggle to provide precise camera control using natural language alone.
In this work, we present a framework for precise camera control with global scene understanding in text-to-image generation by learning parametric camera tokens.
We fine-tune image generation models for viewpoint-conditioned text-to-image generation on a curated dataset that combines 3D-rendered images for geometric supervision and photorealistic augmentations for appearance and background diversity.
Qualitative and quantitative experiments demonstrate that our method achieves state-of-the-art accuracy while preserving image quality and prompt fidelity.
Unlike prior methods that overfit to object-specific appearance correlations, our viewpoint tokens learn factorized geometric representations that transfer to unseen object categories.
Our work shows that text-vision latent spaces can be endowed with explicit 3D camera structure, offering a pathway toward geometrically-aware prompts for text-to-image generation.
Project page: \url{https://randdl.github.io/viewtoken_control/}.
\end{abstract}    
\section{Introduction}
\begin{figure}
\small
    \centering
    \includegraphics[width=1\linewidth]{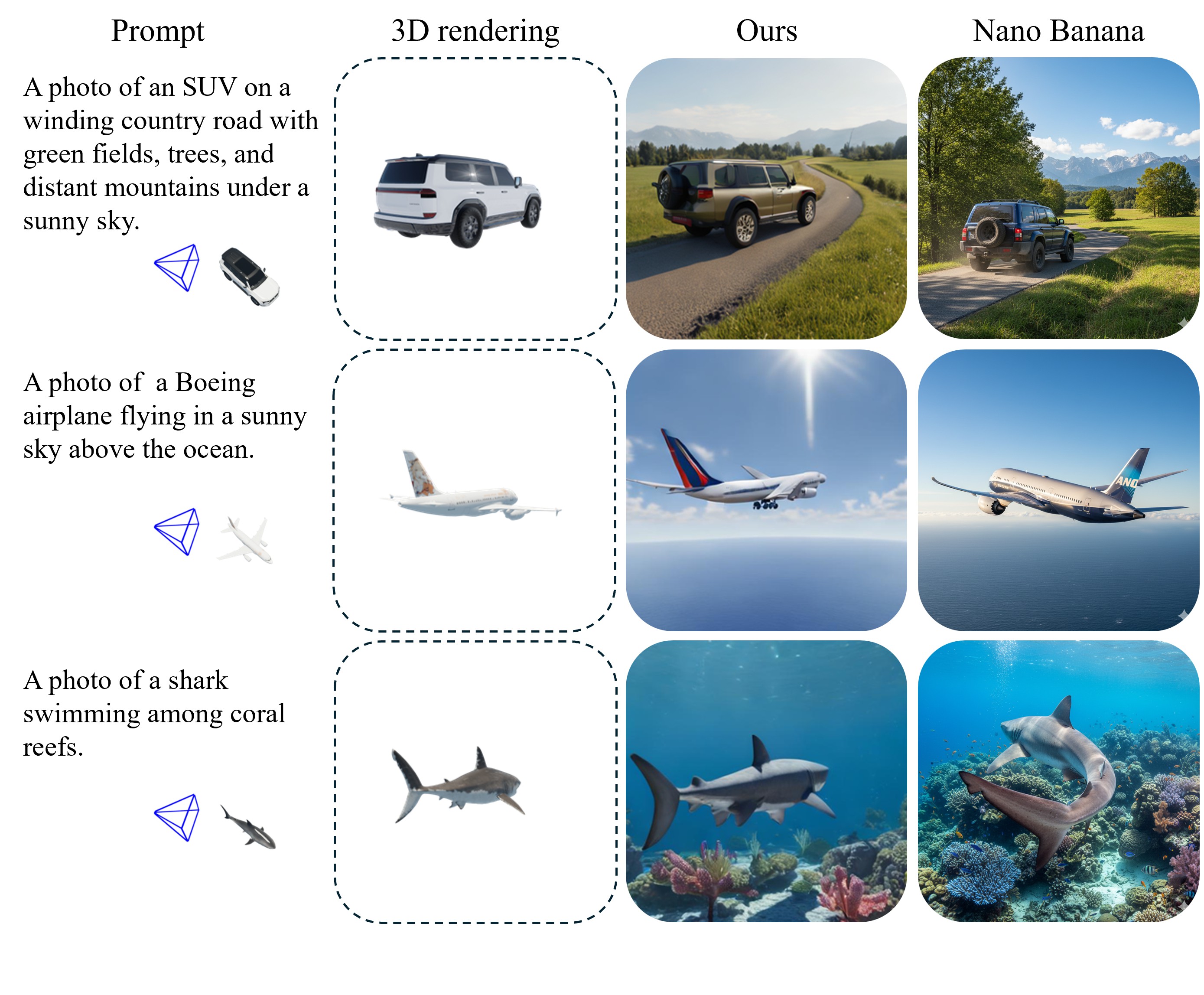}
    \caption{\textbf{Our model vs.\ Gemini 2.5 Flash Image (Nano Banana)}~\cite{Google_Gemini_2.5_Flash_Image}. Our encoded camera viewpoint tokens enable precise camera pose control, while Nano Banana often fails despite detailed descriptions: ``angled diagonally to show its rear and left side from a rear three-quarter view (approx.\ 220° azimuth), with the camera slightly below eye level (10° elevation). It occupies approximately 60\% of the image width, positioned in the center slightly to the left.'' See Supp. \ref{sup:nanobanana_details} for more results. Objects shown in prompts and 3D rendering column are only shown to illustrate the desired viewpoints.}
    \label{fig:motivation}
\end{figure}

Controllable image generation with precise camera viewpoint specification is an increasingly important capability for modern generative models. While many text-to-image models~\cite{rombach2021highresolution, liu2024sorareviewbackgroundtechnology, Google_Gemini_2.5_Flash_Image} have demonstrated remarkable progress in semantic fidelity and visual realism, they struggle to follow even simple geometric instructions such as ``back view'', ``30° left-side view'', or ``45° top-down perspective.'' Natural language is expressive but inherently ambiguous and discrete for viewpoint specification, and current models often hallucinate incorrect poses, collapse to biased canonical angles, or produce inconsistent geometry across trials.
To overcome these limitations, we introduce a method that augments text prompts with explicit, fine-grained camera control, enabling precise specification of viewpoint (\cref{fig:motivation}).


Prior attempts at viewpoint control remain limited. 3D-aware generative models~\cite{tang2023MVDiffusion, shi2023zero123++, long2024wonder3d} and Novel View Synthesis~\cite{liu2023zero, zhou2025stable, qiu2024richdreamer, muller2024multidiff} require additional inputs beyond text prompts as summarized in~\cref{tab:comparison_previous_work}.
View-NeTI~\cite{burgess2025viewpoint} learns object and viewpoint tokens by training on multi-view images for each object. Compass Control~\cite{Parihar_2025_CVPR} makes progress by learning viewpoint tokens with text prompts but only supports azimuth control. Its attention-masking strategy confines viewpoint cross-attention to a local region, which can inhibit global scene understanding and may lead to overfitting to specific training objects.
Overall, learning camera information within text prompts remains underexplored.

\begin{figure}[!t]
\small
\centering
\setlength{\tabcolsep}{1.5pt}
\begin{tabular}{@{}c@{\hspace{1mm}}c@{\hspace{1mm}}c@{}}
\includegraphics[width=0.32\columnwidth]{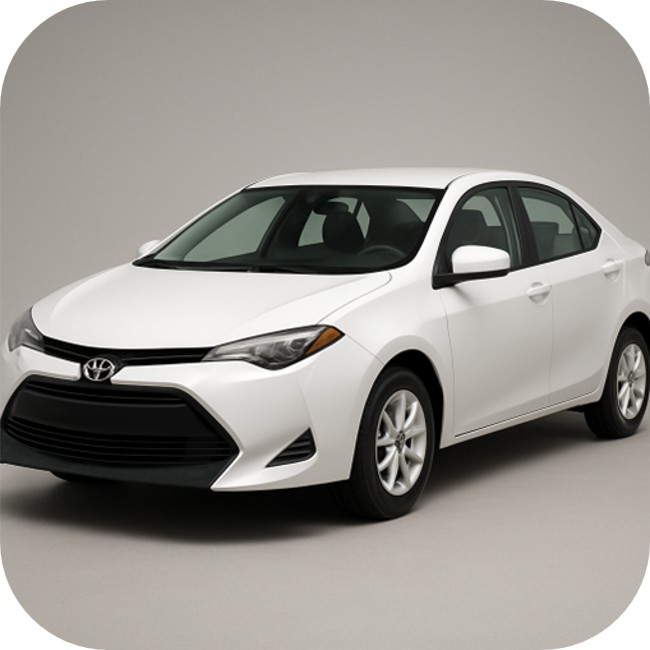} &
\includegraphics[width=0.32\columnwidth]{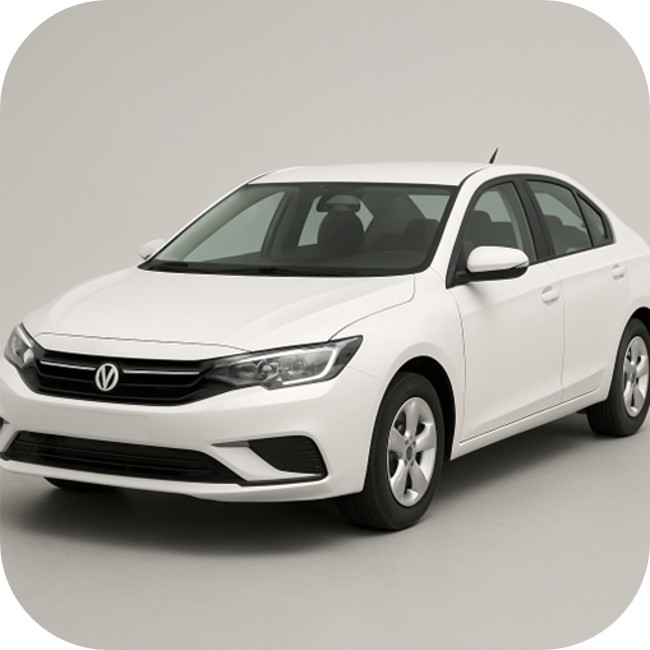} &
\includegraphics[width=0.32\columnwidth]{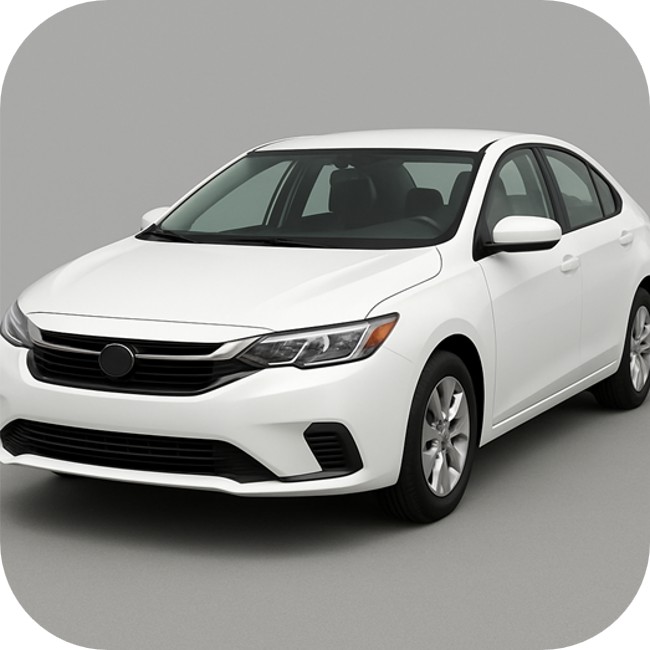} \\
\scriptsize (a) 45° to the left &
\scriptsize (b) 45° to the right &
\scriptsize (c) 30° to the right \\
\end{tabular}
\caption{\textbf{GPT5 viewpoint failures.} Generated by GPT5~\cite{openai2025chatgpt} using ``A white sedan seen from 45°/30° to the left/right of the front view''. All three prompts result in nearly identical orientations.}
\label{fig:gpt}
\end{figure}

\textbf{Approach.}
Our approach adds a camera viewpoint specification to a text prompt and learns how to integrate this information for image generation. The camera is parameterized relative to the object and its front in order to help provide a consistent notion of viewpoint, for example, ``left'' and ``right''. 


These camera parameters are encoded into learnable viewpoint embeddings and concatenated with the text embeddings as the input to a text-to-image generation backbone~\cite{wu2025harmon, rombach2021highresolution, esser2024scaling} and jointly trained or fine-tuned. This allows the resulting trained model to condition on semantic content and explicit camera viewpoint specifications during text-to-image generation.

The choice of data used in this joint training is important to avoid overfitting or collapse. To achieve this we construct a dataset with two parts. One part---the large rendered dataset---uses rendered 3D models of objects. Using this alone for training can cause the text-to-image models to collapse, ``forgetting'' how to render more complex scenes and follow detailed text prompts. To avoid this, we add a second part---photorealistic augmented images---consisting of full scenes containing an object in a known pose. These are generated by prompting a commercial image generation system with a rendered object in a known pose as well as a text description of the object and the scene.  We use the same two-part dataset throughout our experiments.



\begin{figure*}
\small
\centering
\includegraphics[width=0.9\linewidth]{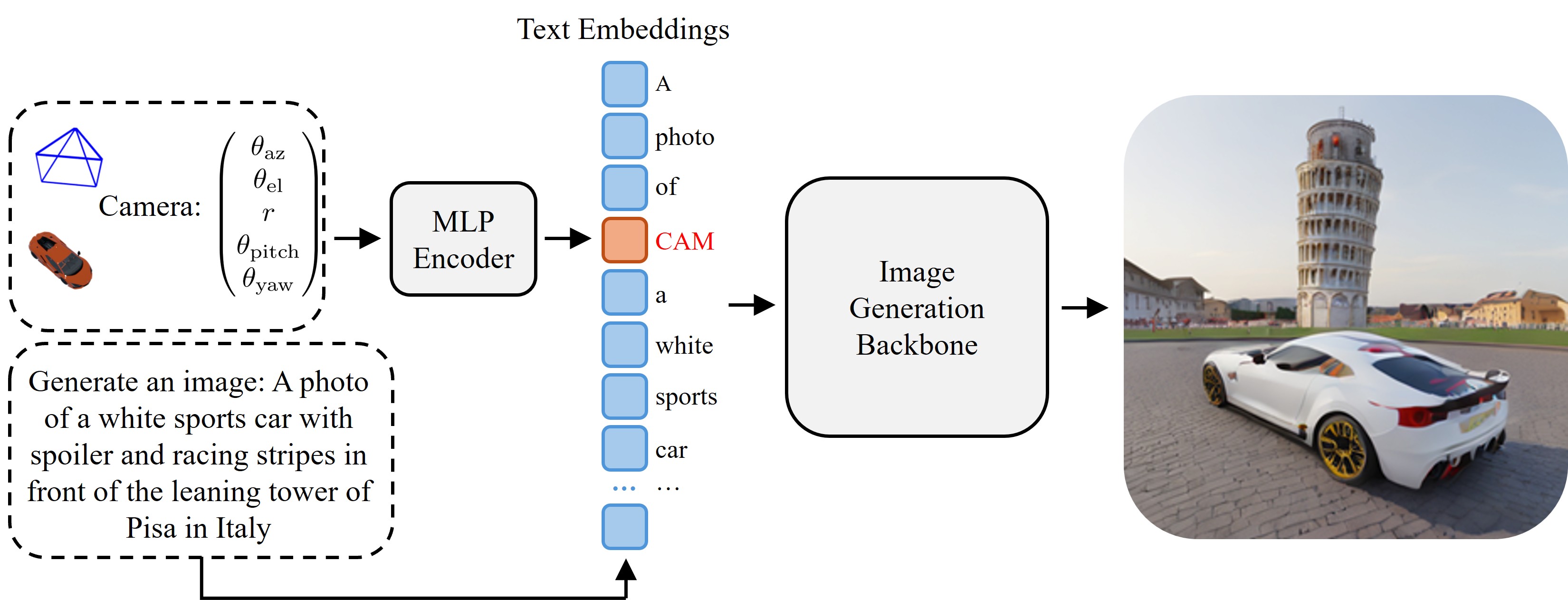}
\caption{\textbf{Architecture overview.} An MLP encoder maps camera parameters to a token embedding that is processed jointly with text tokens for viewpoint-conditioned image generation. We fine-tune the image generation model~\cite{wu2025harmon, rombach2021highresolution, esser2024scaling} jointly with the camera token encoder. The rendered red car in the prompt is only shown to illustrate the desired viewpoint.}
\label{fig:architecture}
\end{figure*}

We demonstrate the effectiveness of our approach by fine-tuning multiple text-to-image generation models, while simultaneously learning a lightweight encoder for each that maps camera parameters to token embeddings (\cref{fig:architecture}).




Through quantitative and qualitative experiments, we show that our method achieves state-of-the-art viewpoint accuracy while maintaining high image fidelity and robust generalization to unseen objects.
Compared to prior work---such as View-NeTI~\cite{burgess2025viewpoint}, which learns object-specific tokens, and Compass Control~\cite{Parihar_2025_CVPR}, which can overfit to training appearance---our method learns viewpoint token embeddings that are more independent of object identity or shape.
Finally, our canonical camera-object framework and two-part dataset design also enable a global understanding of scene geometry, contributing to more consistent foreground–background relationships and more reliable viewpoint control.

\begin{table}[t]
\small
\centering
\caption{\textbf{Comparison with previous work} in terms of input type and camera control.}
\begin{tabular}{lc}
\toprule
\textbf{Model} & \textbf{Input Type} \\
\midrule
Wonder3D~\cite{long2024wonder3d} & Image + Camera \\
Zero-123~\cite{liu2023zero} & Image + Camera \\
OneDiffusion~\cite{le2024diffusiongenerate} & Image + Camera \\
Stable-Virtual-Camera~\cite{zhou2025stable} & Image + Camera \\
Controlnet-Depth~\cite{zhang2023adding} & Image + Depth \\
View-NeTI~\cite{burgess2025viewpoint} & Object Token + Camera Token \\
Compass Control~\cite{Parihar_2025_CVPR} & Text + Azimuth Token \\
\textbf{Ours} & Text + Camera Token \\
\bottomrule
\end{tabular}
\label{tab:comparison_previous_work}
\end{table}

Our contributions are:
\begin{itemize}
    \item State-of-the-art camera control in both range and accuracy while preserving image quality and fidelity to text prompts, outperforming both novel-view generation and prior token-based methods.

    \item Context-preserving training that enables the model to learn viewpoint cues with a global understanding of the scene rather than isolated object cutouts.
    \item Two-part dataset design: a high volume of canonically aligned renderings provide strong geometric supervision, while a low volume of photorealistic augmentations maintain realism and appearance diversity in generations.
    \item Experiments showing that this approach is robust, using the same training data and approach for different text-to-image generation models, and that it has improved generalization to unseen object categories with less overfitting.
\end{itemize}

\section{Related Work}

\textbf{Text-to-Image (T2I) Generation.}
Large-scale diffusion-based text-to-image models~\cite{ramesh2022hierarchical, rombach2021highresolution, song2021scorebased, peebles2023scalable, podell2023sdxl} have achieved unprecedented progress in realism and semantic alignment, while MAR~\cite{li2024autoregressive} provided a strong alternative based on autoregressive image encoding and generation.
More recently, extensions toward unified multimodal models~\cite{wu2025harmon, deng2025bagel, wu2025openuni, xie2024show} further integrate image understanding and generation by learning a unified vision-language space in both input and output.
Despite many advances, such models still struggle to provide precise spatial or geometric control, as natural language offers only implicit viewpoint descriptions, and training data are heavily biased toward front-facing views or common compositions. To overcome these limitations, our approach embeds viewpoint information into the text prompt for better geometry understanding.

\noindent \textbf{Controllable Image Generation.}
To enhance control, many works augment text prompts with auxiliary structural inputs such as depth, edges, or segmentation masks, such as ControlNet~\cite{zhang2023adding} and T2I-Adapter~\cite{mou2023t2i}.
Recent works also explore 2D and 3D layout-guided generation~\cite{li2023gligen, yang2024scenecraft, maillard2025laconic}.
However, methods that condition on structural cues inherently require explicit 3D reference inputs at inference (e.g., depth or edges) thereby limiting flexibility and reducing usability in real-world settings.
In contrast, our method achieves fine-grained spatial control from parameterized camera tokens appended to text inputs without relying on additional geometric information.

\noindent \textbf{3D Generative Models.}
Traditional 3D-aware generative models~\cite{Chan2022, xu20213daware} integrate explicit geometry representations such as NeRF~\cite{mildenhall2020nerf} to synthesize viewpoint-consistent images. Subsequent text-to-3D methods~\cite{poole2022dreamfusion, lin2023magic3d} leverage score distillation sampling to distill pretrained 2D diffusion models into 3D representations.
More recently, many works~\cite{tang2023MVDiffusion, shi2023zero123++, long2024wonder3d, tang2024lgm, shi2023mvdream} directly regress 3D structures from a single or sparse set of images. However, these methods lack a consistent canonical understanding of an object's front-facing orientation, and the resulting 3D models may lack the nuance and realistic texture seen in 2D image generation. In contrast, our work generates viewpoint-conditioned images that preserve high-fidelity appearance and provide contextual consistency between object and the scene background.

\noindent \textbf{Novel View Synthesis.}
Classic NVS methods~\cite{mildenhall2020nerf, kerbl3Dgaussians} reconstruct a scene from multiple calibrated images and enable rendering from unseen viewpoints.
More recent works~\cite{yu2021pixelnerf, wang2021ibrnet, chen2021mvsnerf, liu2023zero, zhou2025stable, qiu2024richdreamer, muller2024multidiff, liu2023syncdreamer, le2024diffusiongenerate} leverage 2D diffusion models to generate multi-view images or 3D structure from one or a few input views.
However, these methods require one or more input images and thus cannot provide direct camera control for \textit{ab initio} text-to-image generation. Our approach bridges this gap by embedding explicit viewpoint tokens into the text prompt, providing a one-step viewpoint-conditioned text-to-image generation model without additional reference images.

\noindent \textbf{Viewpoint-Conditioned Generation.}
Recent work has begun to explore viewpoint control within text-to-image generation models. PreciseCam~\cite{bernal2025precisecam} targets scene-level camera control instead of objects. Diffusion-as-Shader~\cite{gu2025diffusion} addresses relative camera control in video generation rather than T2I. View-NeTI~\cite{burgess2025viewpoint} learns disentangled object and viewpoint tokens but requires object-specific multi-view supervision and struggles to produce geometrically consistent novel views without such data. Compass Control~\cite{Parihar_2025_CVPR} introduces compass tokens that condition generation on azimuth rotations without needing multi-view inputs. 
However, Compass Control's controllability is limited to a single rotation axis and it generalizes poorly to unseen objects. Its attention localization strategy also prevents it from learning a global understanding of the scenes for different viewpoints. Our work extends this line by enabling flexible and accurate camera control over multiple camera parameters and stronger generalization to new objects and prompts.
\section{Method}
Our method is designed to work with any text-to-image model that operates on text embeddings as inputs. As illustrated in \cref{fig:architecture}, given a text prompt and explicit camera parameters $\boldsymbol{\theta}$, we generate a parametric viewpoint embedding token in the same input space as the text tokens. The combined text and viewpoint token embeddings are processed jointly through the model to generate viewpoint-conditioned images. We discuss our object-camera system in \cref{subsec:viewpoint_parameterization}, how we encode the viewpoint into token embeddings in \cref{subsec:viewpoint_mlp}, and our dataset setup in \cref{subsec:dataset}.

\subsection{Viewpoint Parameterization}
\label{subsec:viewpoint_parameterization}
We adopt an object-centric system where the object is fixed at the origin, and the front of the object always faces along the positive x-axis of the world coordinates. This gives us consistent ``left/right'' and ``front/back'' in natural language across all objects. The camera is allowed to move freely to capture different views of the object.

We parameterize the camera viewpoint using a factorized 5-parameter representation:
\begin{equation}
\boldsymbol{\theta} = (\theta_{\text{az}}, \theta_{\text{el}}, r, \theta_{\text{pitch}}, \theta_{\text{yaw}}) \in \mathbb{R}^5
\end{equation}
where $(\theta_{\text{az}}, \theta_{\text{el}}, r)$ defines the position of the camera with a spherical coordinate. The radius $r$ is specified in units of the object diameter. $(\theta_{\text{pitch}}, \theta_{\text{yaw}})$ defines the relative camera rotation with respect to the direction from the camera position to the origin. Positive $\theta_{\text{pitch}}$ represents camera tilting down, and positive $\theta_{\text{yaw}}$ represents camera tilting left. We assume the camera and object ``up'' directions are aligned (i.e., $\theta_{\text{roll}}=0$) and a fixed focal length (FoV of ${\sim}55^\circ$).

\subsection{Viewpoint Token Encoding}
\label{subsec:viewpoint_mlp}

We use a parametric viewpoint token that encodes camera view using a lightweight MLP. Given the 5-parameter viewpoint representation $\boldsymbol{\theta} = (\theta_{\text{az}}, \theta_{\text{el}}, r, \theta_{\text{pitch}}, \theta_{\text{yaw}})$, we first apply a parameter encoding function:
\begin{equation}
\phi(\boldsymbol{\theta}) = [\sin(\theta_{\text{az}}), \cos(\theta_{\text{az}}), \theta_{\text{el}}, r, \theta_{\text{pitch}}, \theta_{\text{yaw}}] \in \mathbb{R}^{6},
\end{equation}
where azimuth is encoded via sine and cosine to handle periodicity, radius is normalized to $[0, 1]$, and elevation, pitch, and yaw are directly used as radian values. We then map these encoded parameters to a token embedding via a 3-layer MLP with ReLU activations:
\begin{equation}
\mathbf{e}_{\text{view}} = \text{MLP}_{\text{view}}(\phi(\boldsymbol{\theta})) \in \mathbb{R}^{d}
\end{equation}
We insert the viewpoint token adjacent to the object description, allowing precise geometric information to flow through the model's attention mechanism alongside text.


\newcolumntype{Z}{>{\centering\arraybackslash}m{0.155\textwidth}}
\newcolumntype{Y}{>{\centering\arraybackslash}m{0.155\textwidth}}

\newcommand{\qualheight}{27mm}

\newcommand{\imgcell}[1]{%
  \includegraphics[height=\qualheight]{qualitative_results/used/#1}%
}

\newcommand{\qualrow}[2]{
  \imgcell{#1_camera} &
  \imgcell{#1_3d} &
  \imgcell{#1_controlnet} &
  \imgcell{#1_novelview} &
  \imgcell{#1_compass} &
  \imgcell{#1_final20} \\[-6pt]
  \multicolumn{6}{l}{\scriptsize \emph{Prompt:} #2} \\[2pt]
}

\begin{figure*}[t]
\small
\centering
\setlength{\tabcolsep}{1.5pt}
\renewcommand{\arraystretch}{1.12}
\begin{tabular}{ZYYYYY}
\toprule
\textbf{Camera Spec} &
\textbf{3D Render \hspace{5cm} (GT View)} &
\textbf{ControlNet} &
\textbf{SV-Camera} &
\textbf{Compass Control} &
\textbf{\;Ours\;} \\
\midrule

\qualrow{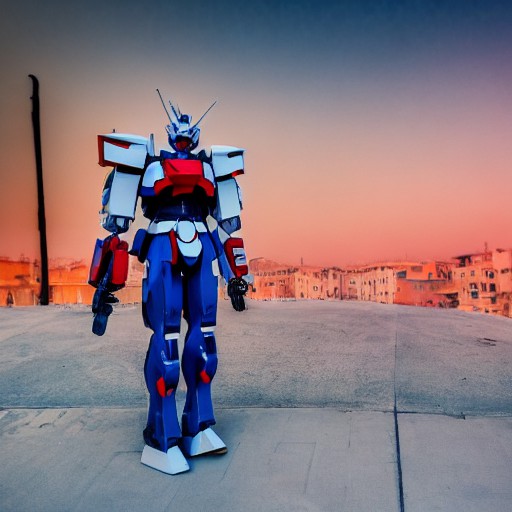}{
  A photo of white gundam robot with blue and red accents on the streets of Venice, with the sun setting in the background
}

\qualrow{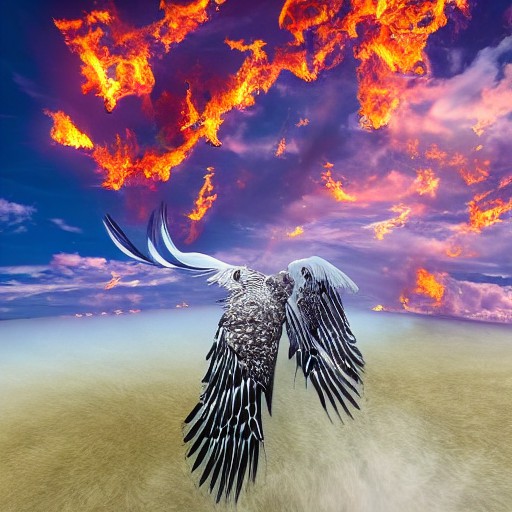}{
  A photo of phoenix rising from flames with wings spread flying high above a sea of fluffy white clouds
}

\qualrow{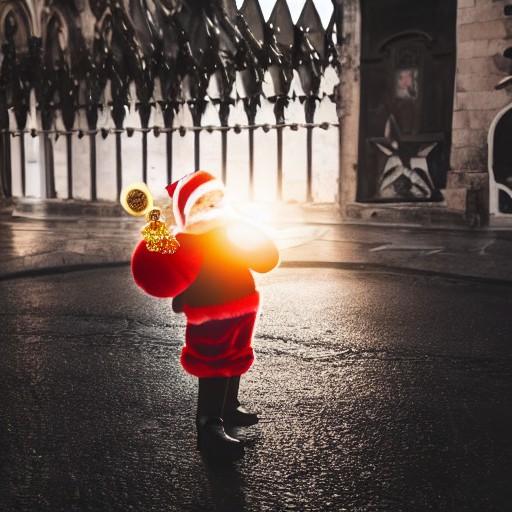}{
  A photo of santa claus with round glasses and black boots on the streets of Venice, with the sun setting in the background
}

\qualrow{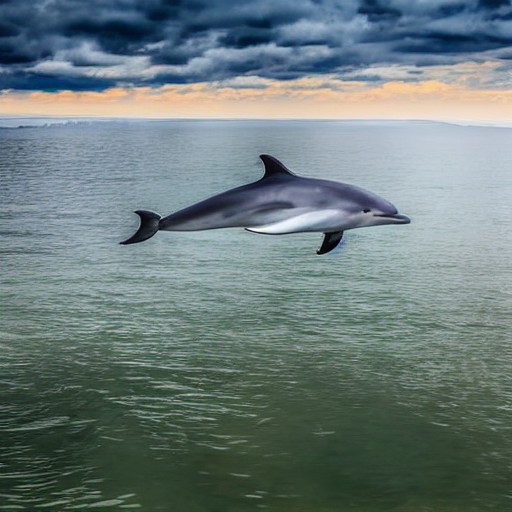}{
  A photo of dolphin on still waters under a cloudy sky, mountains visible in the distant horizon
}

\qualrow{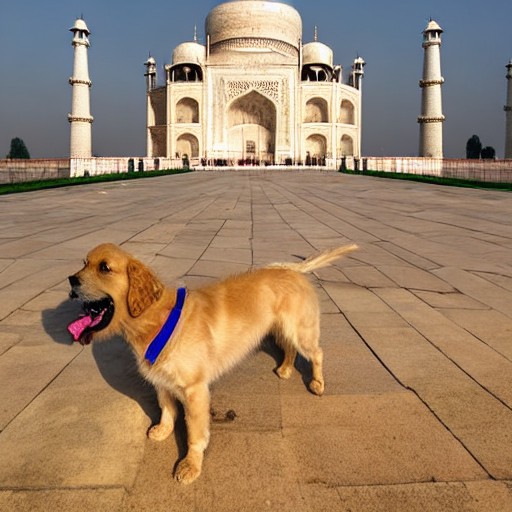}{
  A photo of golden retriever with fluffy fur in front of the Taj Mahal
}


\qualrow{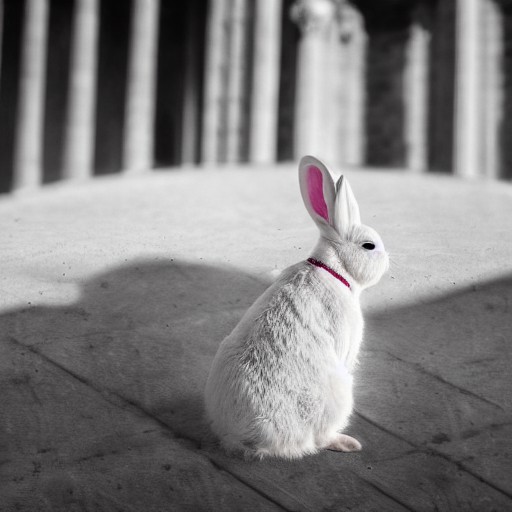}{
  A photo of white bunny with pink ears, holding a carrot in front of the leaning tower of Pisa in Italy
}

\bottomrule
\end{tabular}
\caption{\textbf{Qualitative comparison across methods.} Each row shows images and the corresponding prompt. The first two columns visualize the ground-truth camera frustum and a rendering of a 3D object similar to the prompt description. The depth-map of the rendered object is used as an oracle to guide ControlNet but \textit{not used by the other models}. The remaining columns show results from different methods: ControlNet~\cite{zhang2023adding}, Stable-Virtual-Camera (SV-Camera)~\cite{zhou2025stable}, Compass Control~\cite{Parihar_2025_CVPR}, and our method. Our approach achieves precise viewpoint control while maintaining high image quality and prompt fidelity.}
\label{fig:qualitative_grid}
\end{figure*}

\begin{table*}[h]
\centering
\small
\caption{\textbf{Quantitative comparison} of camera pose fidelity and CLIP score. Methods are evaluated on mean and median angular errors (degrees), radius error (normalized by object size), and CLIP prompt-image similarity. Our approach achieves the best performance across all metrics among models without using oracle geometry information.}
\label{tab:camera_accuracy_main}
\setlength{\tabcolsep}{6pt}
\begin{tabular}{llccccccccc}
\toprule
\multirow{2}{*}{Method} & \multirow{2}{*}{Input Type} &
\multicolumn{2}{c}{Azimuth$\downarrow$} &
\multicolumn{2}{c}{Elevation$\downarrow$} &
\multicolumn{2}{c}{Radius$\downarrow$} &
\multicolumn{1}{c}{Yaw$\downarrow$} &
\multicolumn{1}{c}{Pitch$\downarrow$} &
\multicolumn{1}{c}{CLIP$\uparrow$} \\
\cmidrule(lr){3-4}\cmidrule(lr){5-6}\cmidrule(lr){7-8}\cmidrule(lr){9-9}\cmidrule(lr){10-10}\cmidrule(lr){11-11}
 &  & Mean & Median & Mean & Median & Mean & Median & Mean & Mean & Mean \\
\midrule
ControlNet~\cite{zhang2023adding}      & Image + Oracle Depth & {25.65} & 5.25  & 5.77  & 4.21  & 0.09 & 0.07 & 0.80 & 0.94 & 0.3307 \\
\cmidrule(l{2pt}r{2pt}){1-11}
SV-Camera~\cite{zhou2025stable}        & Image + Camera & 54.89 & 23.52 & 9.05  & 6.99  & 0.20 & 0.18 & 2.89 & 2.29 & 0.2596 \\
Compass~\cite{Parihar_2025_CVPR}       & Text + Azimuth Token & 31.07 & 11.60 & 14.49 & 12.58 & 0.17 & 0.14 & 2.03 & 2.61 & 0.3433 \\
Ours (Harmon)                         & Text + Camera Token  & \textbf{18.11} & \textbf{7.63}  & \textbf{7.62}  & \textbf{6.03}  & \textbf{0.11} & \textbf{0.09} & \textbf{1.25} & \textbf{1.38} & \textbf{0.3555} \\
\bottomrule
\end{tabular}
\end{table*}


\begin{table}[]
\centering
\small
\caption{\textbf{Azimuth error breakdown} across 11 ``easy'' and 26 ``diverse'' objects.}
\label{tab:camera_accuracy_azimuth_breakdown}
\setlength{\tabcolsep}{6pt}
\begin{tabular}{lccc}
\toprule
Method & Whole set & Easy set & Diverse set \\
\midrule
ControlNet~\cite{zhang2023adding}      & 25.65 & 23.22 & 26.87 \\
\cmidrule(l{2pt}r{2pt}){1-4}
SV-Camera~\cite{zhou2025stable} & 54.89 & 62.08 & 51.29 \\
Compass~\cite{Parihar_2025_CVPR}               & 31.07 & {18.62} & 37.29 \\
Ours (Harmon)      & \textbf{18.11} & \textbf{16.22} & \textbf{19.06} \\
\bottomrule
\end{tabular}
\end{table}

\begin{table}[]
\centering
\small
\caption{\textbf{GenEval benchmarks} for single object and color adherence.
}
\label{tab:quality}
\setlength{\tabcolsep}{6pt}
\begin{tabular}{lcc}
\toprule
\multirow{2}{*}{Model} & \multicolumn{2}{c}{GenEval~\cite{ghosh2023geneval} $\uparrow$} \\
\cmidrule(lr){2-3}
 & Single Obj. & Colors \\
\midrule
SD2.1~\cite{rombach2021highresolution}           & 98 & 85 \\
Compass Control~\cite{Parihar_2025_CVPR} & 85.62 & 58.78 \\
\textit{Decrease from SD2.1} & \textit{-14.28} & \textit{-33.82} \\
\cmidrule(l{2pt}r{2pt}){1-3}
Harmon~\cite{wu2025harmon}$^{\dagger}$ & 99.90 & 92.60 \\
Ours (Harmon) & 94.38 & 76.60 \\
\textit{Decrease from Harmon} & \textbf{\textit{-5.52}} & \textbf{\textit{-16.00}} \\
\bottomrule
\end{tabular}
\vspace{2pt}

{\small $^{\dagger}$Using checkpoint from~\cite{xie2025reconstruction}.}
\end{table}


\subsection{Dataset Setup}
\label{subsec:dataset}

For the large rendered dataset, we manually select 3,111 objects across four categories (animals, vehicles, people, and furniture) from TexVerse~\cite{zhang2025texverse}, a large-scale 3D asset collection. We align each object to a canonical front-facing orientation so that $\theta_{\mathrm{az}}=0, \theta_{\mathrm{el}}=0$ correspond to the object's front view. To ensure diverse yet natural perspectives, we sample cameras randomly: $r \in [\frac{4}{3},2]$ object size, $\theta_{\mathrm{az}} \in [0, 2\pi)$, $\theta_{\mathrm{el}} \in [0, \pi/4]$, and $\theta_{\mathrm{pitch}}, \theta_{\mathrm{yaw}} \in [-\pi/12, \pi/12]$.
We render 120 viewpoints per object, yielding around 373K images with transparent background.

To create a second dataset with photorealistic augmented images, we select 800 high-quality objects from the large dataset and render each at 20 random viewpoints for background augmentation. We use Nano Banana~\cite{Google_Gemini_2.5_Flash_Image} to edit the rendered images to include diverse backgrounds and object appearances while following the original rendered pose. We sample a diverse set of detailed descriptions for the objects to encourage better prompt alignment of our method (e.g., ``a horse with a golden body and pale mane'', ``a sports car with sleek body and white racing strips'') and diverse background prompts. We manually filter out implausible results, yielding approximately 6.6K augmented images ($\sim$ 8 viewpoints per object). During training, we sample equally from the rendered and photorealistic datasets.
See Supp. \ref{sup:dataset} for further dataset construction details.

\section{Experiments}
\subsection{Training Details}
We use Harmon~\cite{wu2025harmon}, a unified multimodal model with an LLM backbone~\cite{tang2025qwen25technicalreport} and a MAR decoder~\cite{li2024autoregressive}, as our primary T2I backbone. We fine-tune the backbone jointly with the viewpoint MLP using the backbone's standard image generation loss.
We initialize from a pretrained Harmon checkpoint~\cite{xie2025reconstruction} and fine-tune for 7{,}500 iterations with a batch size of 192 using AdamW~\cite{loshchilov2017decoupled}.
We apply separate learning rates: a higher rate of $2\times10^{-4}$ for the newly introduced ViewpointMLP and a lower rate of $2\times10^{-5}$ for the pretrained Harmon LLM and MAR decoder.
Training takes approximately 28 hours on a single NVIDIA A100 (80~GB).

\subsection{Evaluation Metrics}
We evaluate our method on two aspects: (i) viewpoint accuracy for each camera parameter, and (ii) prompt alignment using CLIP similarity~\cite{radford2021learning} and the GenEval benchmark~\cite{ghosh2023geneval}.

\textbf{Viewpoint Accuracy.}
To measure geometric fidelity, we train a viewpoint regressor following similar evaluation protocol of Compass Control~\cite{Parihar_2025_CVPR}. The regressor achieves a mean azimuth error of $4.16^\circ$ on images synthesized with ControlNet and Canny edges of rendered 3D objects.
We provide further details of the regressor in Supp. \ref{sup:regressor}.

\begin{table*}[]
\centering
\small
\caption{\textbf{Challenging viewpoints.} Camera pose accuracy and CLIP score. Values in parentheses show differences compared to the main test set.}
\label{tab:camera_accuracy_comparison_hard}
\setlength{\tabcolsep}{6pt}
\begin{tabular}{lcccccccccc}
\toprule
\multirow{2}{*}{Method} &
\multicolumn{3}{c}{Azimuth$\downarrow$} &
\multirow{2}{*}{Elevation$\downarrow$} &
\multirow{2}{*}{Radius$\downarrow$} &
\multirow{2}{*}{Yaw$\downarrow$} &
\multirow{2}{*}{Pitch$\downarrow$} &
\multirow{2}{*}{CLIP$\uparrow$} \\
\cmidrule(lr){2-4}
 & Whole & Easy & Diverse &  &  &  &  &  \\
\midrule
ControlNet~\cite{zhang2023adding} & 33.74{\scriptsize(+8.09)} & 30.31{\scriptsize(+7.09)} & 35.45{\scriptsize(+8.58)} & 5.51{\scriptsize(-0.26)} & 0.10{\scriptsize(+0.01)} & 0.77{\scriptsize(-0.03)} & 1.92{\scriptsize(+0.98)} & 0.3162{\scriptsize(-0.0145)} \\
\cmidrule(l{2pt}r{2pt}){1-9}
Compass~\cite{Parihar_2025_CVPR}           & 39.07{\scriptsize(+8.00)} & 28.66{\scriptsize(+10.04)} & 44.27{\scriptsize(+6.98)} & 21.21{\scriptsize(+6.72)} & 0.19{\scriptsize(+0.02)} & 1.77{\scriptsize(-0.26)} & 4.79{\scriptsize(+2.18)} & {0.3421}{\scriptsize(-0.0012)} \\
Ours (Harmon) & \textbf{23.27}{\scriptsize(+5.16)} & \textbf{21.56}{\scriptsize(+5.34)} & \textbf{24.11}{\scriptsize(+5.05)} & \textbf{9.82}{\scriptsize(+2.20)} & \textbf{0.12}{\scriptsize(+0.01)} & \textbf{1.27}{\scriptsize(+0.02)} & \textbf{2.57}{\scriptsize(+1.19)} & \textbf{0.3522}{\scriptsize(-0.0033)} \\
\bottomrule
\end{tabular}
\end{table*}

\textbf{Prompt Alignment.}
To verify that viewpoint conditioning preserves the model’s text-to-image generation ability, we evaluate on general benchmarks. CLIP similarity quantifies semantic correspondence between generated images and prompts, while the \textit{Single Object} and \textit{Color} cases in GenEval assess object presence and descriptive fidelity.

\subsection{Testing Dataset}
We evaluate on 11 ``easy'' test objects from Compass Control~\cite{Parihar_2025_CVPR} and 26 additional ``diverse'' objects spanning animals, vehicles, furniture, people, and mythical creatures. Eleven of these objects do not appear in our training data, testing cross-category generalization.
For each ``diverse'' object, we generate three descriptive phrases and combine them with background prompts to test the generalization abilities. Each object-background pair is rendered with 10 random viewpoints, totaling 5{,}550 test samples.

To test robustness and object-background consistency on challenging camera angles, we construct an additional test set with 2 back views ($\theta_{\text{az}} \in [\tfrac{3}{4}\pi, \tfrac{5}{4}\pi]$)
and 2 high-elevation views ($\theta_{\text{el}} = \tfrac{2\pi}{9}$) per object–background combination, totaling 2,220 samples.

\subsection{Baselines}
\begin{itemize}
    \item \textbf{ControlNet-Depth}~\cite{zhang2023adding} (text + depth). Provides an oracle baseline with perfect geometry by using depth maps from 3D objects placed on a ground plane.
    \item \textbf{Stable-Virtual-Camera (SV-Camera)}~\cite{zhou2025stable} (image + camera). Performs novel-view synthesis given a front-view image input, which we generate using ControlNet-Plus~\cite{controlnetplus} with depth map conditioning.
    \item \textbf{Compass Control}~\cite{Parihar_2025_CVPR} (text + object azimuth). Encodes azimuth orientation tokens and uses 2D bounding boxes for localization. For fair comparison, we extract 2D boxes from test object renderings as additional conditions.
\end{itemize}

\subsection{Quantitative Results}
\textbf{Viewpoint Accuracy.}
\Cref{tab:camera_accuracy_main} summarizes quantitative results. Our method has lower errors than Compass Control and Stable-Virtual-Camera across all five camera parameters. ControlNet-Depth performs slightly better on a few parameters due to oracle access to depth information.
\Cref{tab:camera_accuracy_azimuth_breakdown} shows a breakdown of the azimuth error on the ``easy'' set and the ``diverse'' set. Compass Control shows a large discrepancy between performance on the 11 ``easy’’ and 26 ``diverse’’ objects, while our method maintains low azimuth errors on both sets, demonstrating stronger generalization.

\noindent\textbf{Prompt Alignment.}
As shown in \cref{tab:quality,tab:camera_accuracy_main}, our method achieves higher GenEval scores than Compass Control and a higher CLIP similarity than all baselines. Our approach maintains better prompt fidelity of the backbone model than Compass Control does. Compass Control often fails to generate the correct object or color specified in the prompt, reflecting overfitting to its training distribution.

\noindent\textbf{Challenging Viewpoints.}
\Cref{tab:camera_accuracy_comparison_hard} reports performance on challenging back-view and high-elevation configurations. Our method retains superior accuracy under these extreme conditions, whereas Compass Control degrades sharply. ControlNet-Depth achieves high pitch accuracy due to explicit geometric supervision, but its text alignment remains weaker. These results demonstrate our method’s robustness to rare and difficult viewpoints.

\noindent\textbf{Generalization Ability.}
We quantify the overfitting problem of Compass Control by calculating the percentage of overfitting to its training objects. Among the three testing objects (``Santa Claus'', ``dolphin'', and ``rabbit''), Compass Control overfits to lions, ostriches, shoes, sofas, and teddy bears 94.2\% of the time, showing that it overfits to category-specific correlations rather than learning factorized viewpoint representations.
In contrast, our method has no obvious overfitting among the three objects and consistently produces semantically correct outputs across all categories, indicating effective disentanglement between viewpoint and object identity.
See Supp. \ref{sup:compass} for a detailed report.

\subsection{Qualitative Results}
\begin{figure}[h]
\small
\centering
\setlength{\tabcolsep}{1.5pt}
\begin{tabular}{@{}c@{\hspace{1mm}}c@{\hspace{1mm}}c@{\hspace{1mm}}c@{}}
\includegraphics[width=0.24\columnwidth]{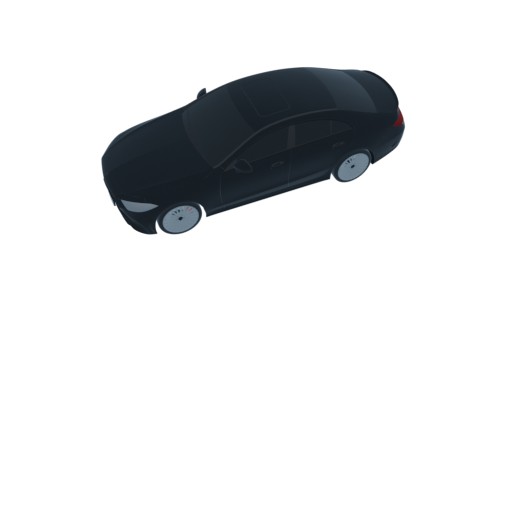} &
\includegraphics[width=0.24\columnwidth]{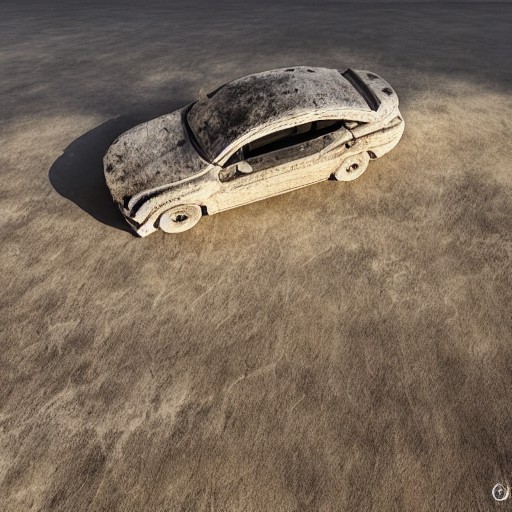} &
\includegraphics[width=0.24\columnwidth]{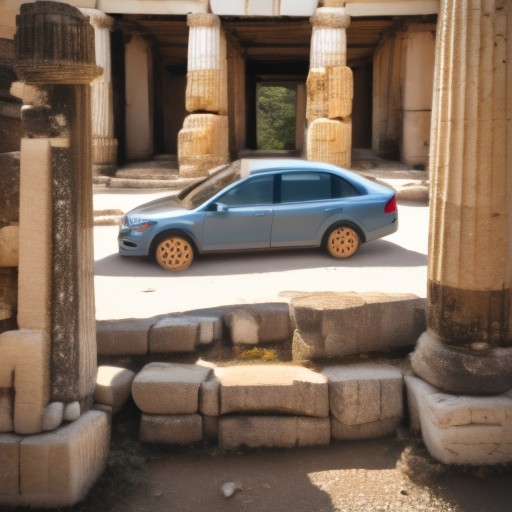} &
\includegraphics[width=0.24\columnwidth]{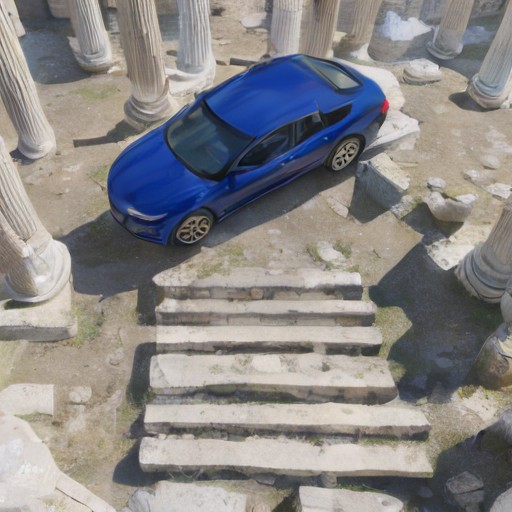} \\
\scriptsize (a) Rendered &
\scriptsize (b) ControlNet &
\scriptsize (c) Compass &
\scriptsize (d) Ours \\
\end{tabular}
\caption{\textbf{Comparisons on a high-angle view.} Prompt: A photo of a sedan in an ancient Greek temple ruin, with broken columns and weathered stone steps.}
\label{fig:high_angle}
\end{figure}

\begin{figure}[h]
\small
\centering
\setlength{\tabcolsep}{1.5pt}
\begin{tabular}{@{}c@{\hspace{1mm}}c@{\hspace{1mm}}c@{}}
\includegraphics[width=0.32\columnwidth]{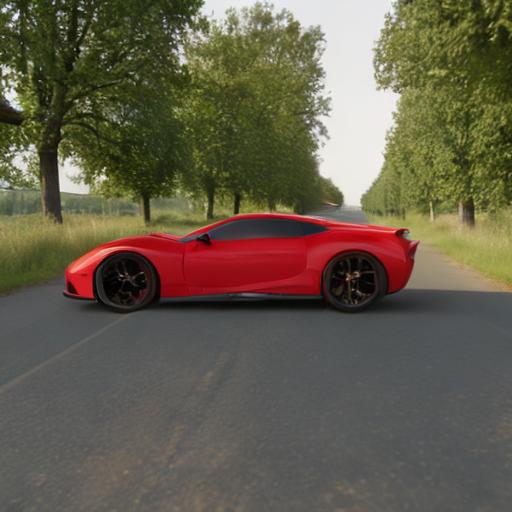} &
\includegraphics[width=0.32\columnwidth]{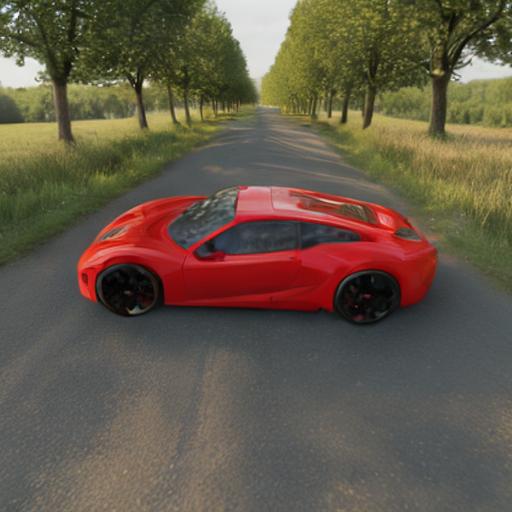} &
\includegraphics[width=0.32\columnwidth]{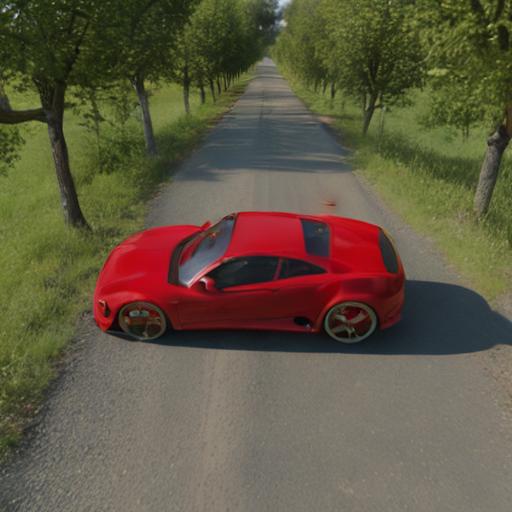} \\
\scriptsize (a) 0° &
\scriptsize (b) 20° &
\scriptsize (c) 40° \\
\end{tabular}
\caption{\textbf{Results at varying camera elevations}: 0, 20, 40 degrees. The horizon line changes with the camera elevation.}
\label{fig:elevations}
\end{figure}

\Cref{fig:qualitative_grid} compares methods qualitatively across diverse camera viewpoints and text prompts. ControlNet preserves object contours via access to accurate geometry but often fits geometrically incoherent content into shapes without semantic awareness. Stable-Virtual-Camera struggles with occluded regions, sometimes producing invalid shapes for novel viewpoints. Compass Control reproduces correct azimuths for seen categories but overfits heavily for novel objects—e.g., generating ``Santa Claus'' as an animal, ``dolphin'' as a four-legged creature, ``rabbit'' as a teddy bear.
In contrast, our method generalizes well across novel categories (e.g., Gundam, Phoenix, Santa Claus), demonstrating that our viewpoint tokens capture better geometric conditions independent of object semantics.

High-elevation cases further highlight this distinction: other methods fail to follow the prompt or the viewpoint. This limitation arises from Compass Control’s restricted cross-attention mechanism to local object regions and T2I backbone's training bias towards eye-level viewpoints. However, our method learns viewpoint conditioning jointly across foreground and background, producing globally coherent compositions, as shown in \Cref{fig:high_angle,fig:elevations}. This global understanding demonstrates the potential of learning 3D geometry information in text prompts for image generation.

We demonstrate our method's robustness and generalization further on objects that do not exist in reality. \Cref{fig:non-existent} presents three examples generated from imaginative text prompts. Our method produces visually plausible and diverse images that faithfully follow both the prompt semantics and the specified viewpoints. It further shows that we preserve the backbone models' understanding of text and images while embedding viewpoint tokens into the input space for text-to-image generation.

To demonstrate the extensibility of our framework, we retrain a variant of our method on the Compass Control two-object dataset. As shown in \cref{fig:two_objects}, it allows independent control over each object’s orientation.

\subsection{Ablations and Variations}


\begin{table}[]
\centering
\small
\caption{\textbf{Backbone Variation and Ablation study}}
\label{tab:ablation}
\setlength{\tabcolsep}{2pt} 
\begin{tabular}{lccccc}
\toprule
Method & Azimuth & Elevation & Radius & Yaw & Pitch \\
\midrule
Ours (Harmon) & 18.11 & 7.62 & 0.11 & 1.25 & 1.38 \\
Ours (SD2.1) & 19.16 & 6.82 & 0.11 & 1.18 & 1.16 \\
Ours (SD3.5) & 12.85 & 8.09 & 0.17 & 2.75 & 1.97 \\
\midrule
Pl\"ucker rays & 21.61 & 8.43 & 0.19 & 1.29 & 1.51 \\
12D matrix & 24.44 & 8.74 & 0.20 & 4.89 & 4.66 \\
sinusoidal encoding & 60.90 & 9.05 & 0.13 & 1.69 & 1.78 \\
\midrule
No rendered subset & 22.98 & 9.34 & 0.20 & 4.84 & 4.93 \\
Freezing backbone & 40.19 & 8.47 & 0.14 & 1.83 & 2.07 \\
More tokens & 18.03 & 7.45 & 0.13 & 1.80 & 1.87 \\
\bottomrule
\end{tabular}
\end{table}
The following experiments are summarized in \Cref{tab:ablation}.
\noindent\textbf{Backbone.}
To isolate the contribution of our method from the Harmon backbone, we train two variants using Stable Diffusion 2.1~\cite{rombach2021highresolution} and Stable Diffusion 3.5~\cite{esser2024scaling} with the same camera encoding architecture. They both achieve comparable viewpoint accuracy, confirming that the geometric generalization stems from our method and dataset rather than the backbone alone.

\noindent\textbf{Viewpoint Encoding.}
We compare our factorized encoding against Pl\"ucker rays, 12D camera matrices, and sinusoidal positional encodings, which all underperform our encoding. We hypothesize that while Pl\"ucker rays excel at dense pixel-wise correspondence for multi-view generation and sinusoidal positional encodings capture high-frequency details, our single-view T2I setting benefits from disentangled semantic orientation signals that are easier for the model to learn. High-frequency components will introduce training instability and the entangled representation will make it difficult to separate camera position (azimuth and elevation) from camera rotation (yaw and pitch).

\noindent\textbf{Other Ablations.}
Removing the rendered subset of the training data leads to a substantial accuracy drop, confirming its importance for learning geometric consistency. Fine-tuning both the LLM and MAR modules is also crucial, suggesting the backbone model~\cite{wu2025harmon} does not have 3D geometry-aware representations in its text input space. Adding additional tokens yields no observable benefit.

\begin{figure}[h]
\small
\centering
\setlength{\tabcolsep}{1.5pt}
\begin{tabular}{@{}c@{\hspace{1mm}}c@{\hspace{1mm}}c@{}}
\includegraphics[width=0.32\columnwidth]{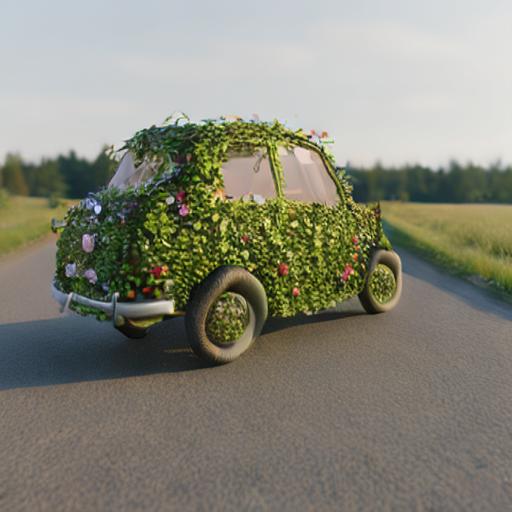} &
\includegraphics[width=0.32\columnwidth]{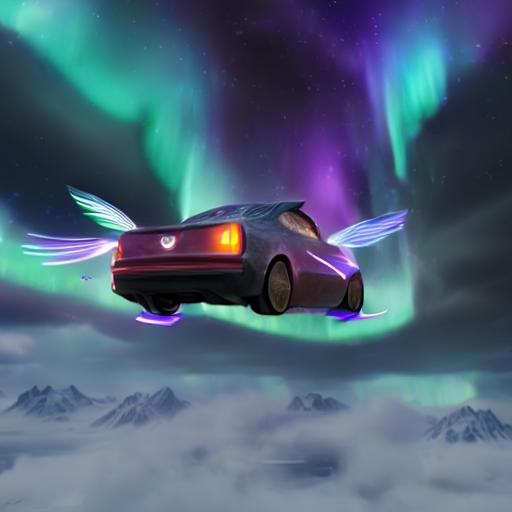} &
\includegraphics[width=0.32\columnwidth]{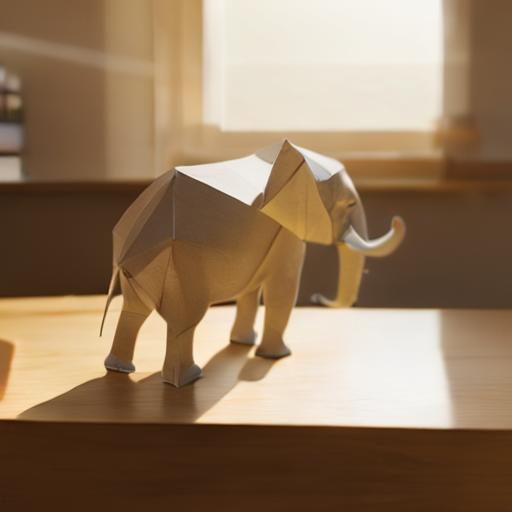} \\
\scriptsize (a) &
\scriptsize (b) &
\scriptsize (c) \\
\end{tabular}
\caption{\textbf{Non-existent objects.} They use the same viewpoint as \cref{fig:motivation}. (a): A small car made of vines and flowers on a countryside road, (b): A flying car with wings made of energy ribbons flying through a storm of glowing auroras over the Arctic, (c): An origami elephant standing on a wooden desk under soft sunlight.}
\label{fig:non-existent}
\end{figure}

\begin{figure}[h]
\small
\centering
\setlength{\tabcolsep}{1.5pt}
\begin{tabular}{@{}c@{\hspace{1mm}}c@{\hspace{1mm}}c@{}}
\includegraphics[width=0.32\columnwidth]{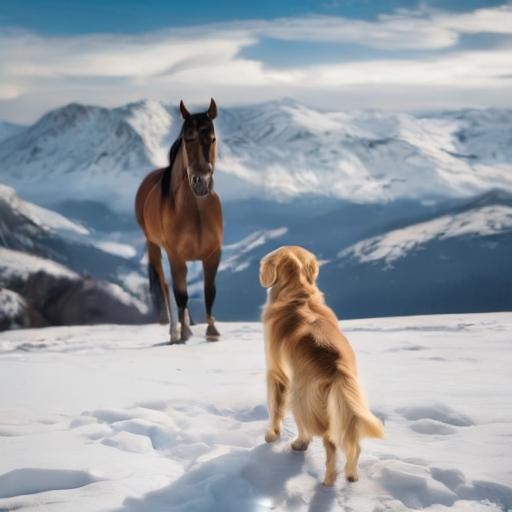} &
\includegraphics[width=0.32\columnwidth]{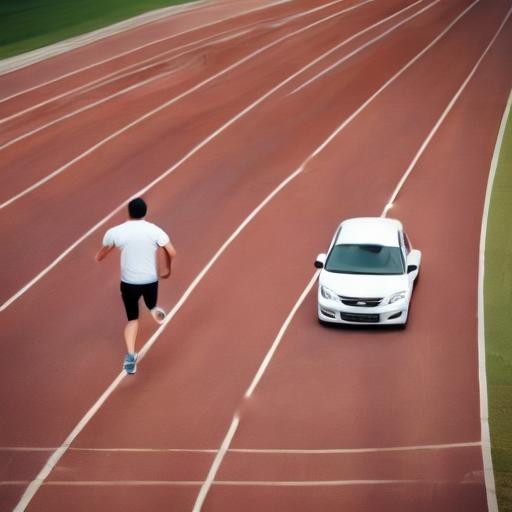} &
\includegraphics[width=0.32\columnwidth]{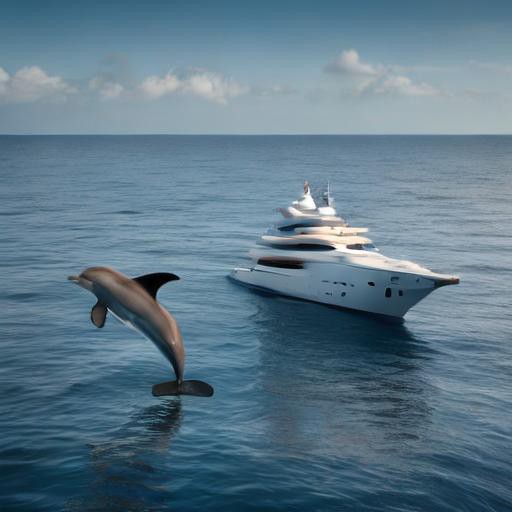} \\
\scriptsize (a) &
\scriptsize (b) &
\scriptsize (c) \\
\end{tabular}
\caption{\textbf{Multi-object viewpoint control.} (a) Golden retriever and horse, azimuth: 170°, -10°. (b) Running man and sedan, azimuth: -160°, 20°. (c) Dolphin and yacht, azimuth: 120°, -60°.}
\label{fig:two_objects}
\end{figure}

\begin{figure}[h]
\small
\centering
\setlength{\tabcolsep}{1.5pt}
\begin{tabular}{@{}c@{\hspace{1mm}}c@{\hspace{1mm}}c@{}}
\includegraphics[width=0.32\columnwidth]{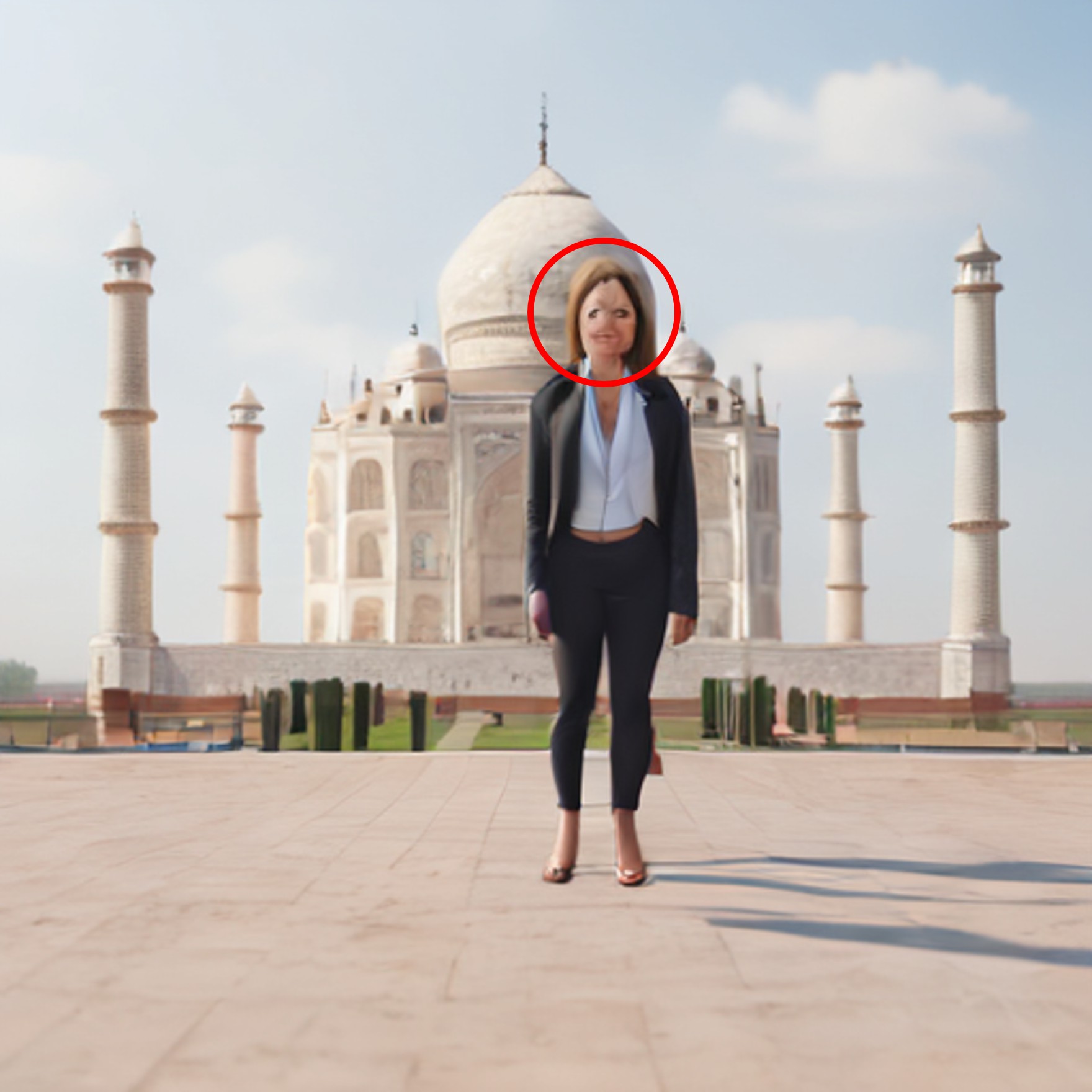} &
\includegraphics[width=0.32\columnwidth]{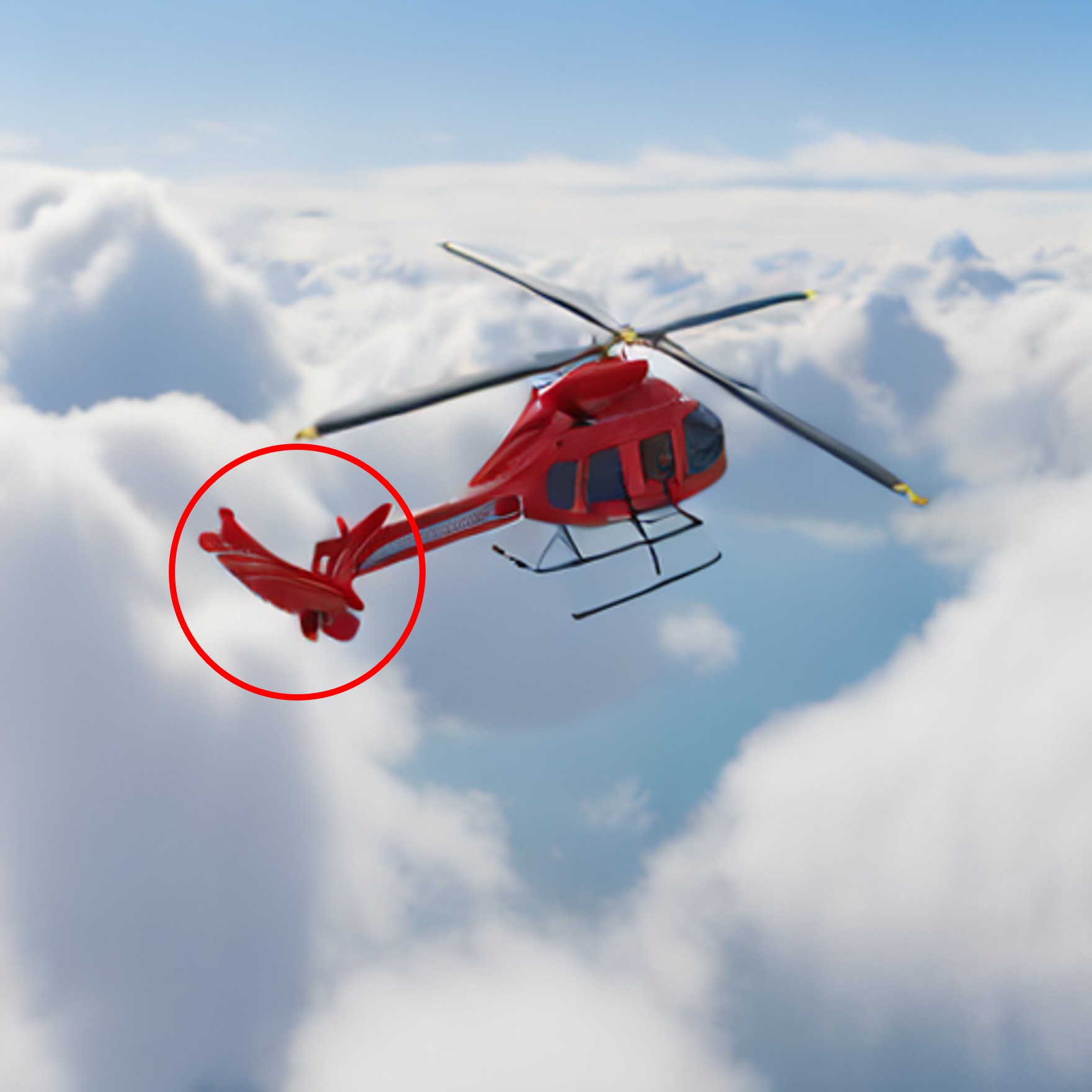} &
\includegraphics[width=0.32\columnwidth]{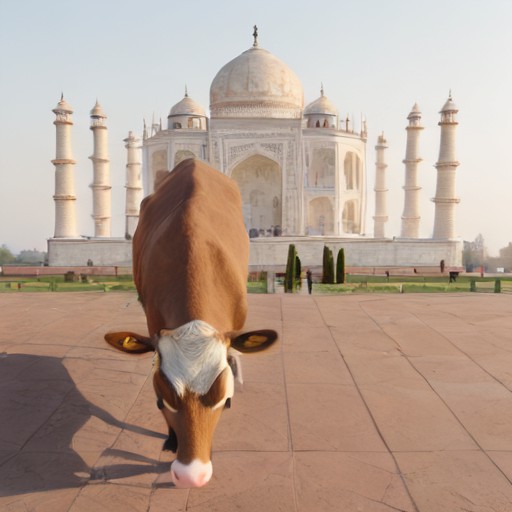} \\
\scriptsize (a) &
\scriptsize (b) &
\scriptsize (c) \\
\end{tabular}
\caption{\textbf{Examples of failure cases.} (a–b) Red circles highlight errors; (c) Misaligned background viewpoints.}
\label{fig:failure}
\end{figure}

\begin{figure}[t]
\small
\centering
\setlength{\tabcolsep}{1.5pt}
\begin{tabular}{@{}c@{\hspace{1mm}}c@{\hspace{1mm}}c@{\hspace{1mm}}c@{}}
\includegraphics[width=0.24\columnwidth]{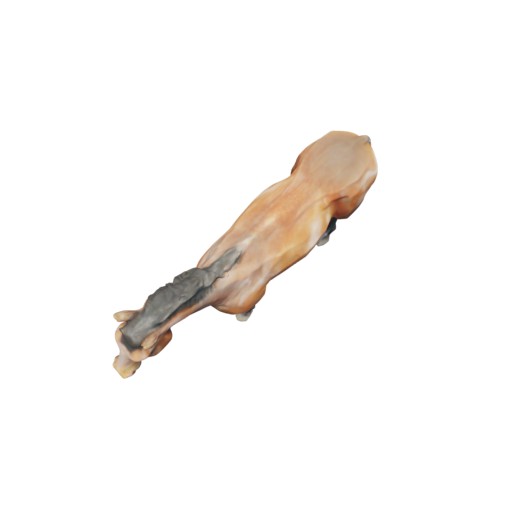} &
\includegraphics[width=0.24\columnwidth]{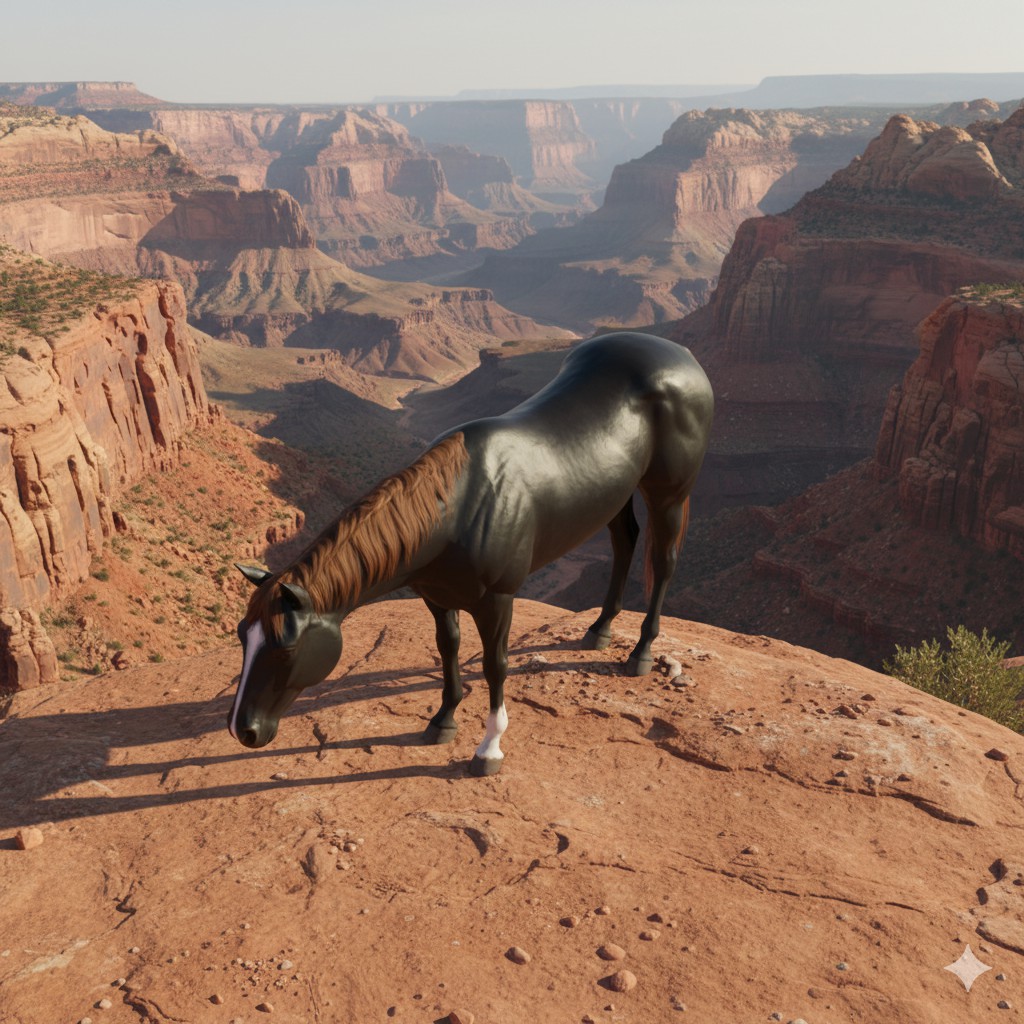} &
\includegraphics[width=0.24\columnwidth]{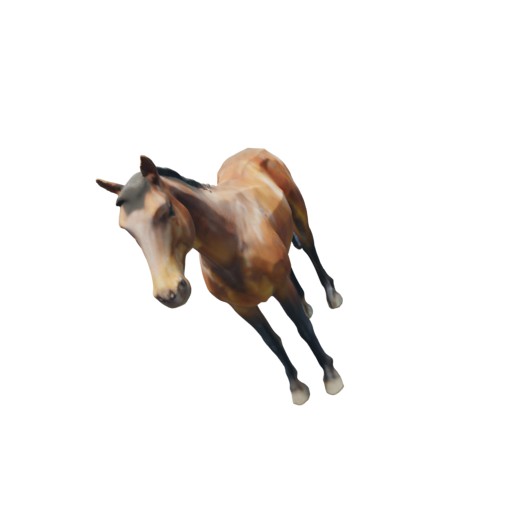} &
\includegraphics[width=0.24\columnwidth]{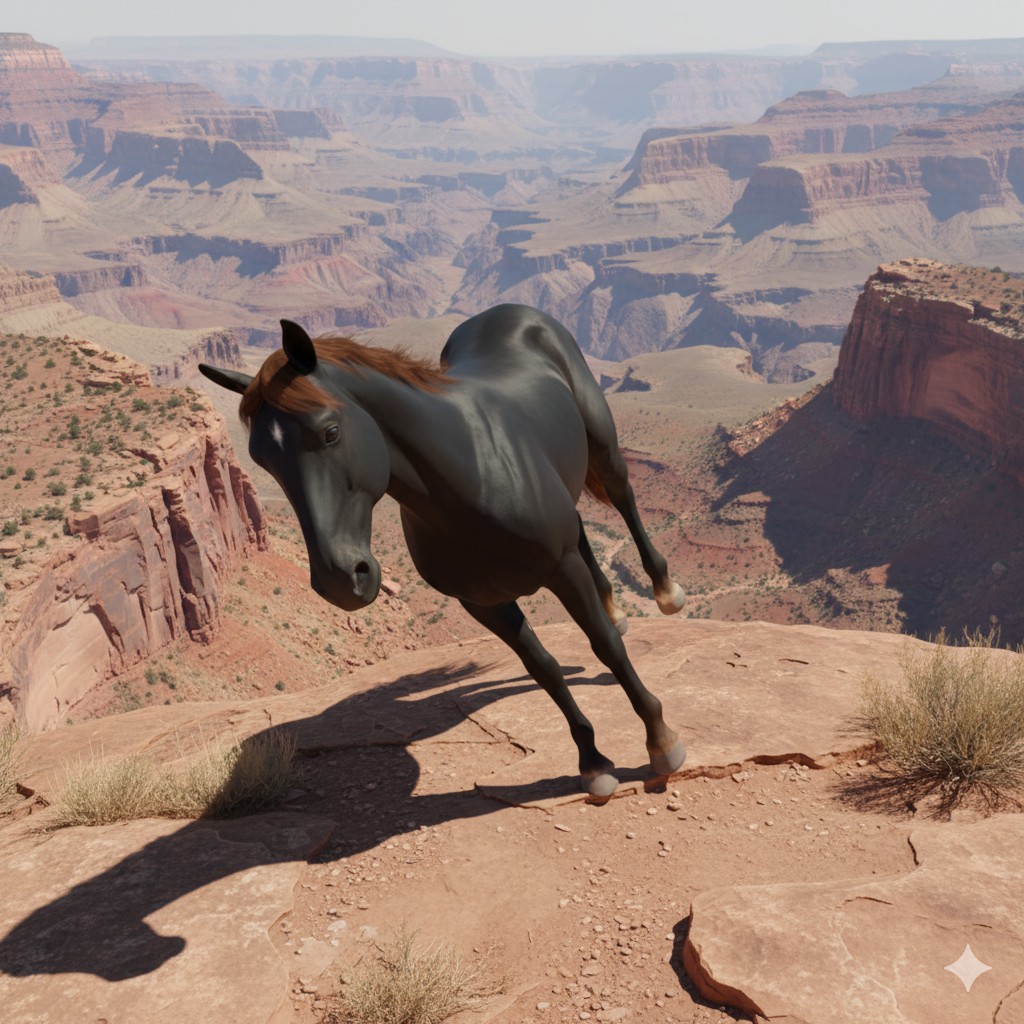} \\
\scriptsize (a) &
\scriptsize (b) &
\scriptsize (c) &
\scriptsize (d) \\
\end{tabular}
\caption{Examples of \textbf{Nano Banana}~\cite{Google_Gemini_2.5_Flash_Image} dataset augmentation failing at extreme elevation ($75^\circ$, a--b) and roll ($30^\circ$, c--d). (a, c) Reference; (b, d) augmented output.}
\label{fig:viewpoint_selection}
\end{figure}

\subsection{Limitations}
Even though our method is capable of generating diverse viewpoints, the T2I backbones have a strong prior toward eye-level, horizontally centered views, particularly for well-known landmarks (e.g., ``Taj Mahal''). This bias can cause the model to favor centered horizons for prompts with landmarks.
We also occasionally observe degraded generations in human faces and fine structural details.
\Cref{fig:failure} illustrates the three representative failure cases.
Our dataset currently covers elevation angles only in $[0^\circ, 45^\circ]$ and excludes roll rotations and intrinsics, as synthesizing reliable photorealistic data for rare viewpoints remains challenging. Even with ground-truth 3D renderings as reference, Nano Banana~\cite{Google_Gemini_2.5_Flash_Image} often fails under extreme viewpoints (\cref{fig:viewpoint_selection}).

\section{Conclusion}
We present a method for precise camera viewpoint control in text-to-image generation through learnable viewpoint tokens.
By fine-tuning an image generation model on curated 3D renderings with photorealistic augmentation, we achieve state-of-the-art viewpoint accuracy while preserving image quality and prompt fidelity.
Compared to previous work, our approach expands from azimuth control to flexible camera control and encourages global scene understanding, which is important for generating images with a consistent background and object viewpoint.
Our method and dataset design work for different backbones and promote generalization to unseen object categories.
Overall, results demonstrate that text prompts can internalize explicit 3D camera structure through simple parametric encoding, opening a new pathway toward geometrically-aware text-to-image generation systems.

\textbf{Acknowledgements.} This work was supported in part by the DARPA Perceptually enabled Task Guidance (PTG) Program under contract number HR00112220005, and by funding from the UCI CS Department.\\
{
    \small
    \bibliographystyle{ieeenat_fullname}
    \bibliography{main}
}

\renewcommand{\thesection}{\Alph{section}}
\renewcommand{\theHsection}{Supplement.\Alph{section}} 
\addcontentsline{toc}{section}{Supplementary Material} 
\clearpage
\setcounter{page}{1}
\setcounter{section}{0}
\maketitlesupplementary

\section{Code and Training Details}
Our project is implemented using Python and PyTorch. We build much of the implementation upon the source code released by Harmon~\cite{wu2025harmon}.
Training follows a cosine-annealed schedule with 1\% linear warmup and gradient clipping (norm 1.0). We use an MLP of 3 layers with a hidden dimension of 1024 and an output dimension the same as the token embeddings.

\section{More Nano Banana Results}
\label{sup:nanobanana_details}
\begin{figure*}
\small
    \centering
    \includegraphics[width=1\linewidth]{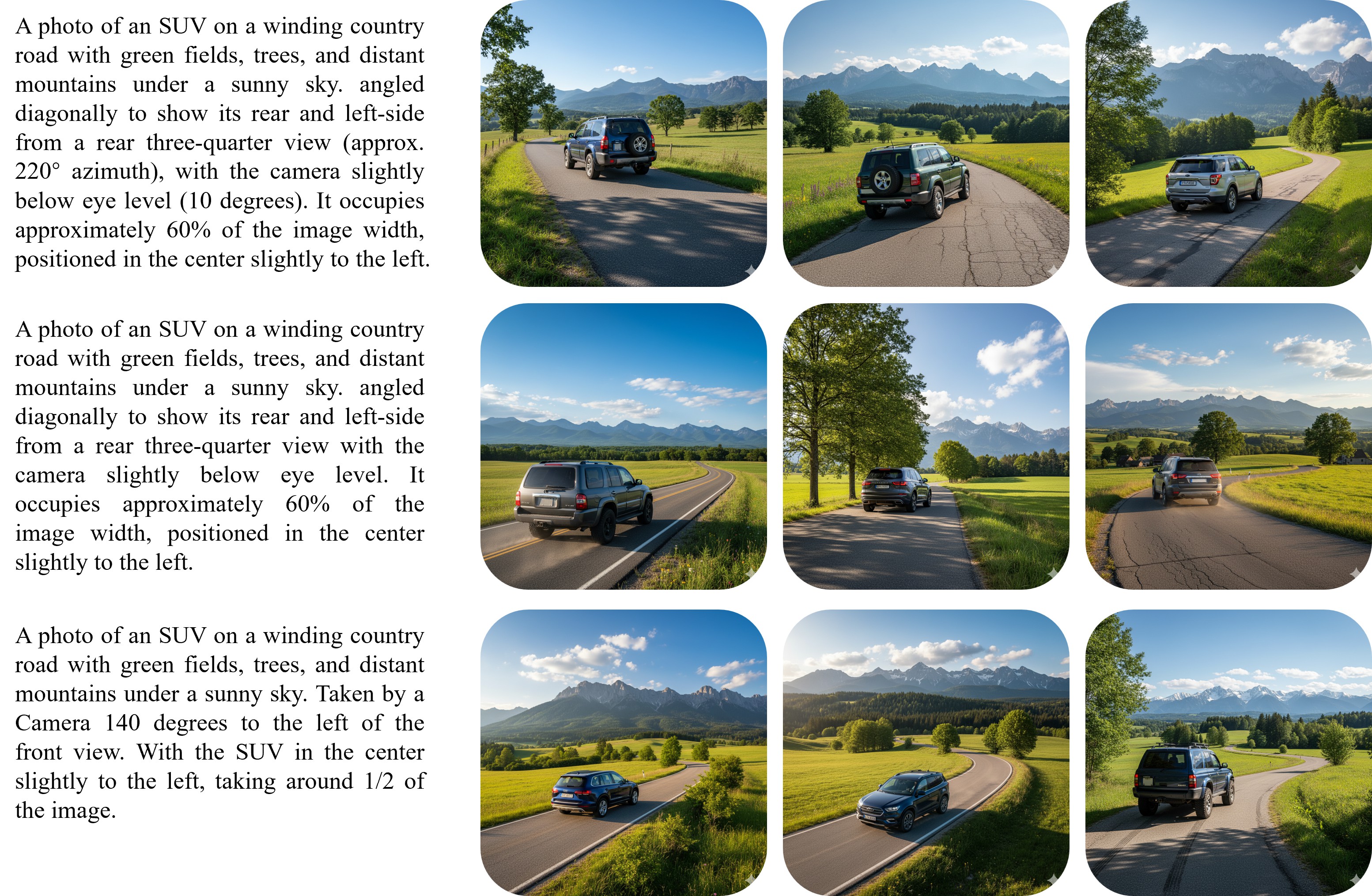}
    \caption{\textbf{More Nano Banana results} with different camera prompts~\cite{Google_Gemini_2.5_Flash_Image}.}
    \label{fig:nanobanana}
\end{figure*}

\Cref{fig:nanobanana} shows more results from Nano Banana with different prompts we tried. We ask Gemini to describe the camera position of the 3D rendering to generate the first two camera prompts. We write the third camera prompt.

\newcommand{\compheight}{27mm}

\newcommand{\compimg}[1]{%
  \includegraphics[height=\compheight]{qualitative_results/compass_overfitting/#1}%
}

\newcommand{\compcell}[2]{
  \begin{minipage}{0.49\textwidth}
    \centering
    \compimg{#1_3d} \hspace{1pt}
    \compimg{#1_compass} \hspace{1pt}
    \compimg{#1_final20} \\[2pt]
    \scriptsize \emph{Object: #2}
  \end{minipage}%
}

\begin{figure*}[t]
\small
\centering

\begin{minipage}{0.49\textwidth}
  \centering
  \scriptsize
  \makebox[0.32\textwidth]{\textbf{3D Rendering}} \hspace{1pt}
  \makebox[0.32\textwidth]{\textbf{Compass Control}} \hspace{1pt}
  \makebox[0.32\textwidth]{\textbf{Ours}}
\end{minipage}
\hspace{4pt}
\begin{minipage}{0.49\textwidth}
  \centering
  \scriptsize
  \makebox[0.32\textwidth]{\textbf{3D Rendering}} \hspace{1pt}
  \makebox[0.32\textwidth]{\textbf{Compass Control}} \hspace{1pt}
  \makebox[0.32\textwidth]{\textbf{Ours}}
\end{minipage}

\vspace{2pt}

\compcell{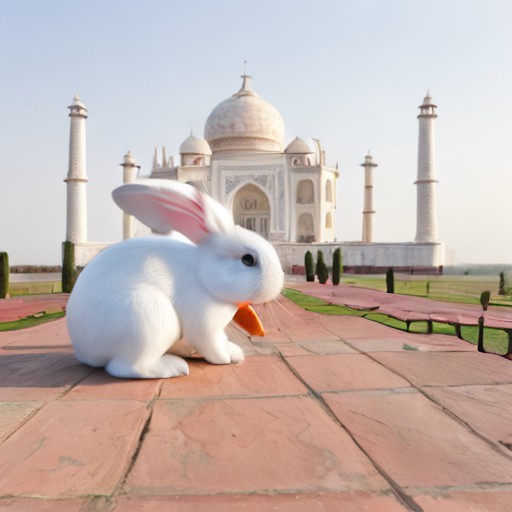}{
  white bunny with pink ears, holding a carrot
}
\hspace{2pt}
\compcell{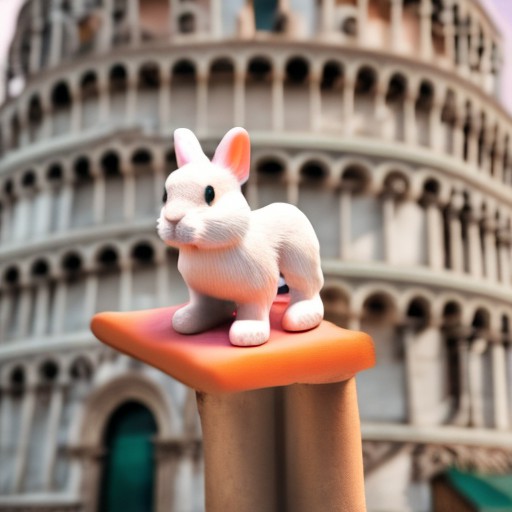}{
  white bunny with pink ears, holding a carrot
}

\vspace{8pt}

\compcell{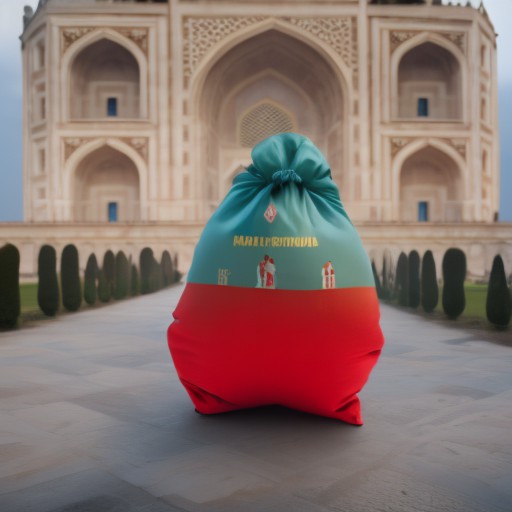}{
  santa claus carrying a sack of gifts
}
\hspace{2pt}
\compcell{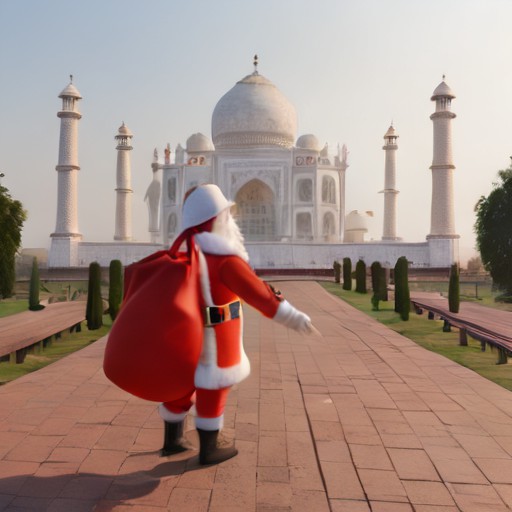}{
  santa claus carrying a sack of gifts
}

\vspace{8pt}

\compcell{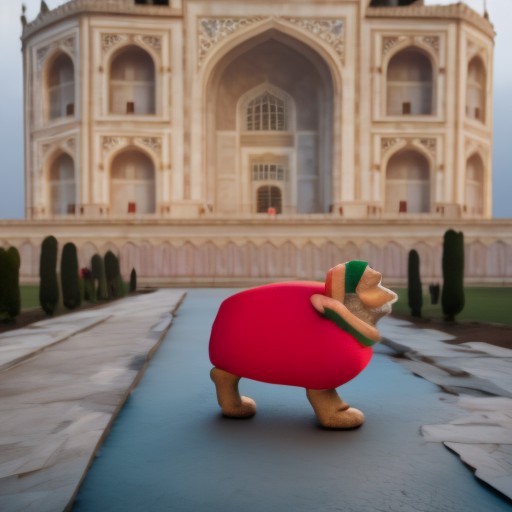}{
  santa claus carrying a sack of gifts
}
\hspace{2pt}
\compcell{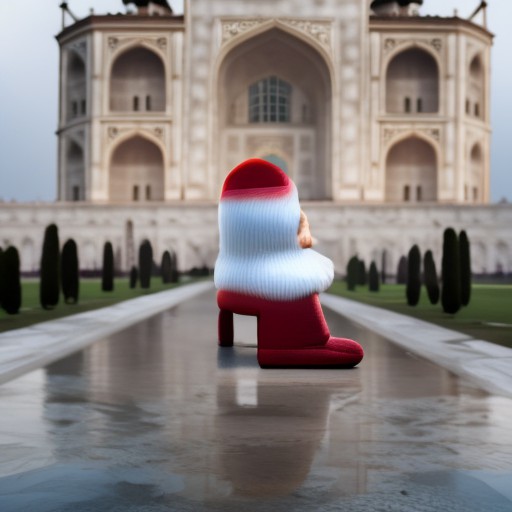}{
  santa claus with round glasses and black boots
}

\vspace{4pt}

\caption{\textbf{Compass Control vs.\ ours.} Each example shows three images: \textbf{Left:} 3D ground truth rendering, \textbf{Middle:} Compass Control~\cite{Parihar_2025_CVPR}, \textbf{Right:} Our method. The comparison demonstrates our model's improved viewpoint control and generalization compared to Compass Control.}
\label{fig:overfitting_compass}
\end{figure*}

\begin{figure}[]
\centering
\begin{tikzpicture}
\pie[
  radius=2.3,
  text=legend,
  line width=0.4pt
]{
  8.67/{Correct (39)},
  56.22/{Lion (253)},
  14.89/{Ostrich (67)},
  3.78/{Teddy bear (17)},
  10.44/{Unknown (47)},
  2.89/{Shoe (13)},
  3.11/{Sofa (14)}
}
\end{tikzpicture}
\caption{\textbf{Compass Control overfitting distribution.} Rendered images for novel test objects categorized as ``correct'', similar to a training object, or ``unknown'' (150 images each for Santa Claus, rabbit, and dolphin).}
\label{fig:compass_pie}
\end{figure}
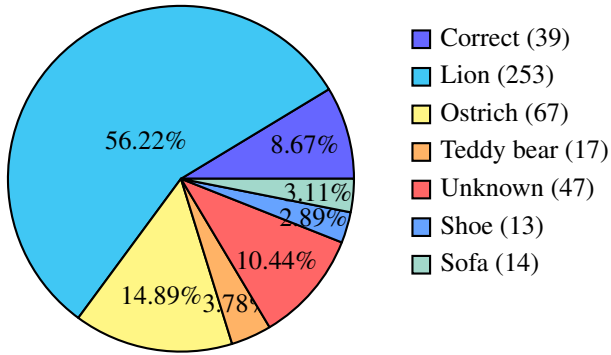

\section{Training Dataset}\label{sup:dataset}

\textbf{Camera Settings}
We use a focal length of 35mm and Blender's default sensor size of 36 mm, resulting in an FOV of 54.4.

\noindent \textbf{Object selection and normalization.}
We only include objects with semantically unambiguous front-facing orientations (e.g., the front of a vehicle, the face of an animal, the interactive side of furniture) and normalize the scale of each object to fit in a square bounding box of side length 1.


\noindent \textbf{Rendered Dataset.}
The full set of 3,111 objects is rendered at 120 random viewpoints each against transparent backgrounds, providing dense viewpoint coverage (373,320 total images). We generate captions for all objects to enable text-conditioned generation.

\noindent \textbf{Photorealistic Augmented Dataset.}
From the 3{,}111 objects, we select 800 diverse, highest-quality assets for photorealistic augmentation. Each object is rendered from 20 random viewpoints and processed through Gemini 2.5 Flash Image~\cite{Google_Gemini_2.5_Flash_Image} model with the rendered image and an image editing prompt: \emph{``Using the provided image, maintain the \{object\_name\}'s location, pose, and head/gaze direction; remove all 3D rendering cues, polygon edges, and flat surfaces; transform the \{object\_name\} into a new photorealistic \{object\_name\} with the following NEW features: \{desc\_text\}; and inpaint the transparent background with \{background\} so that the \{object\_name\} appears organically integrated into the scene with correct relative size, lighting, shadows, atmospheric perspective, and natural interaction with the environment.''}

\begin{figure}[t]
\centering
\setlength{\tabcolsep}{1.5pt}
\begin{tabular}{@{}c@{\hspace{2mm}}c@{\hspace{2mm}}c@{}}
\includegraphics[width=0.31\columnwidth]{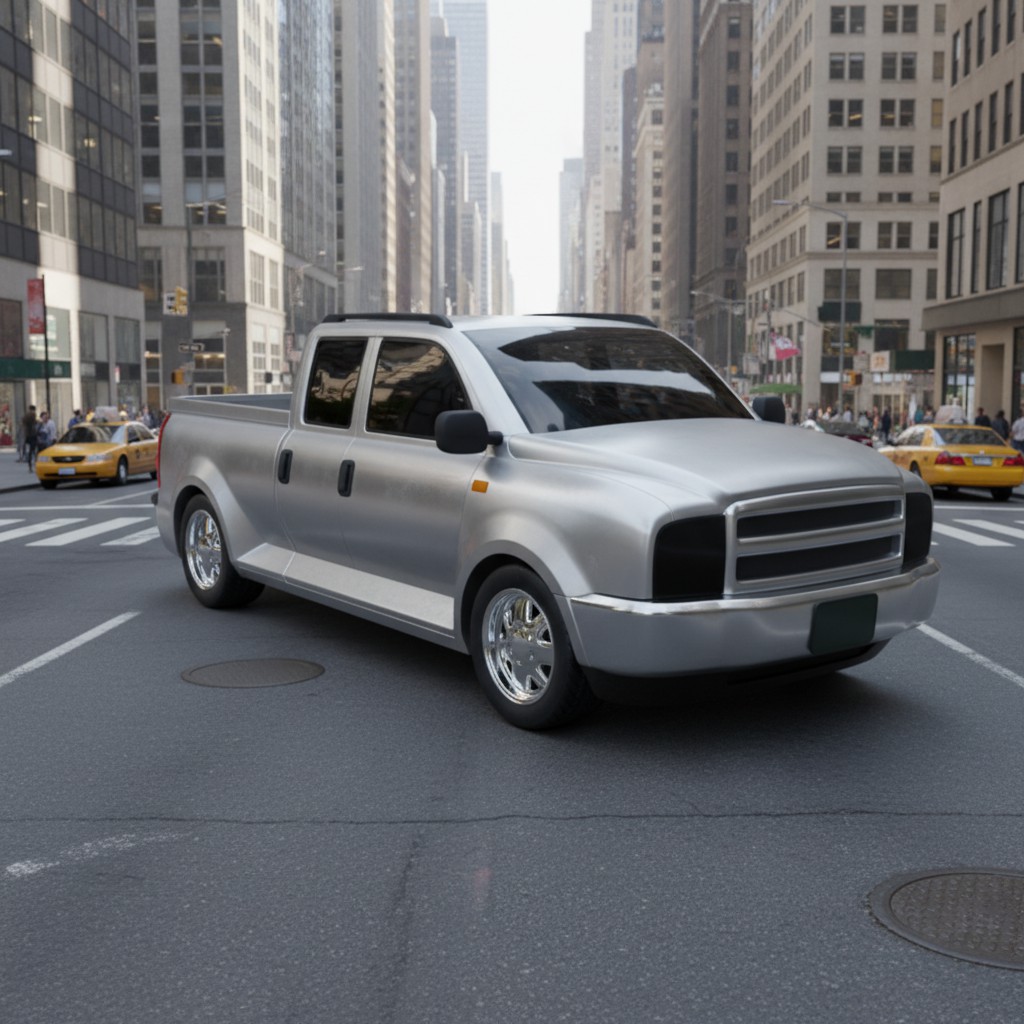} &
\includegraphics[width=0.31\columnwidth]{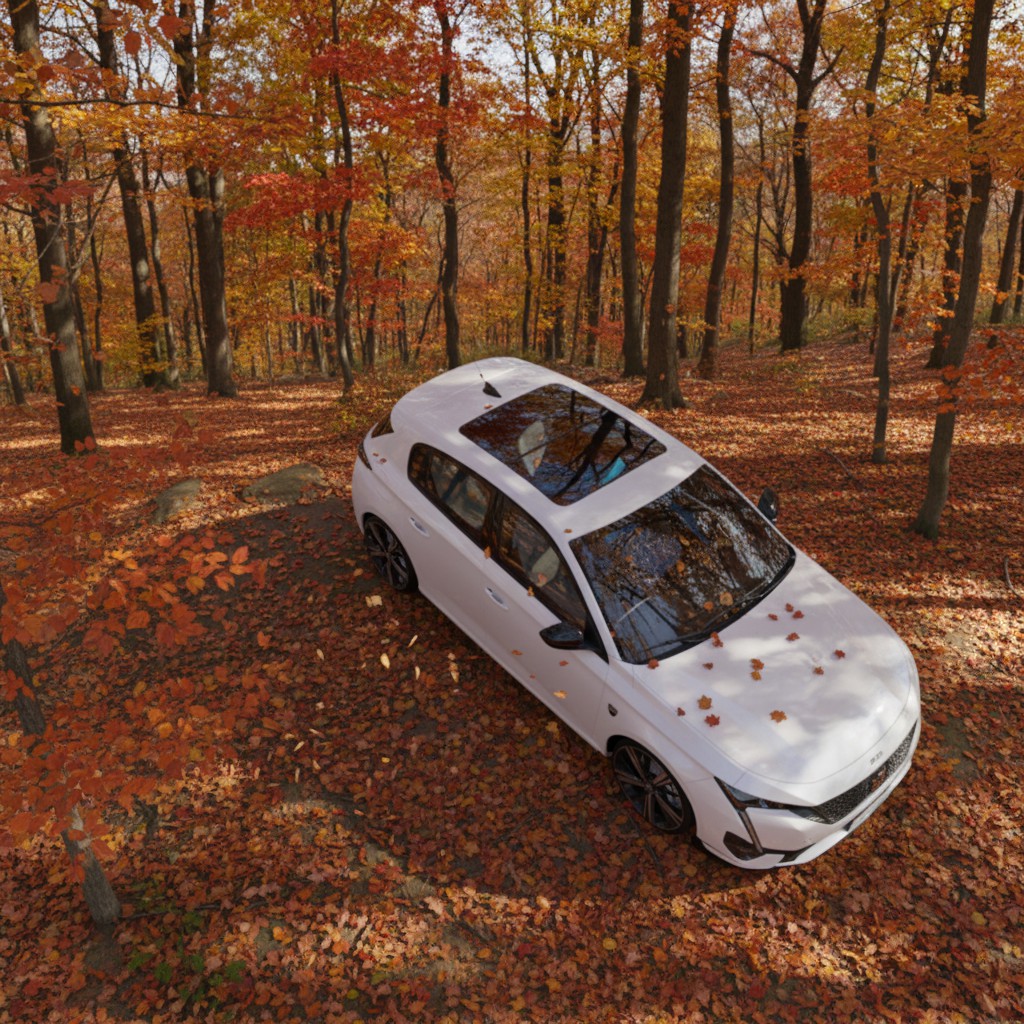} &
\includegraphics[width=0.31\columnwidth]{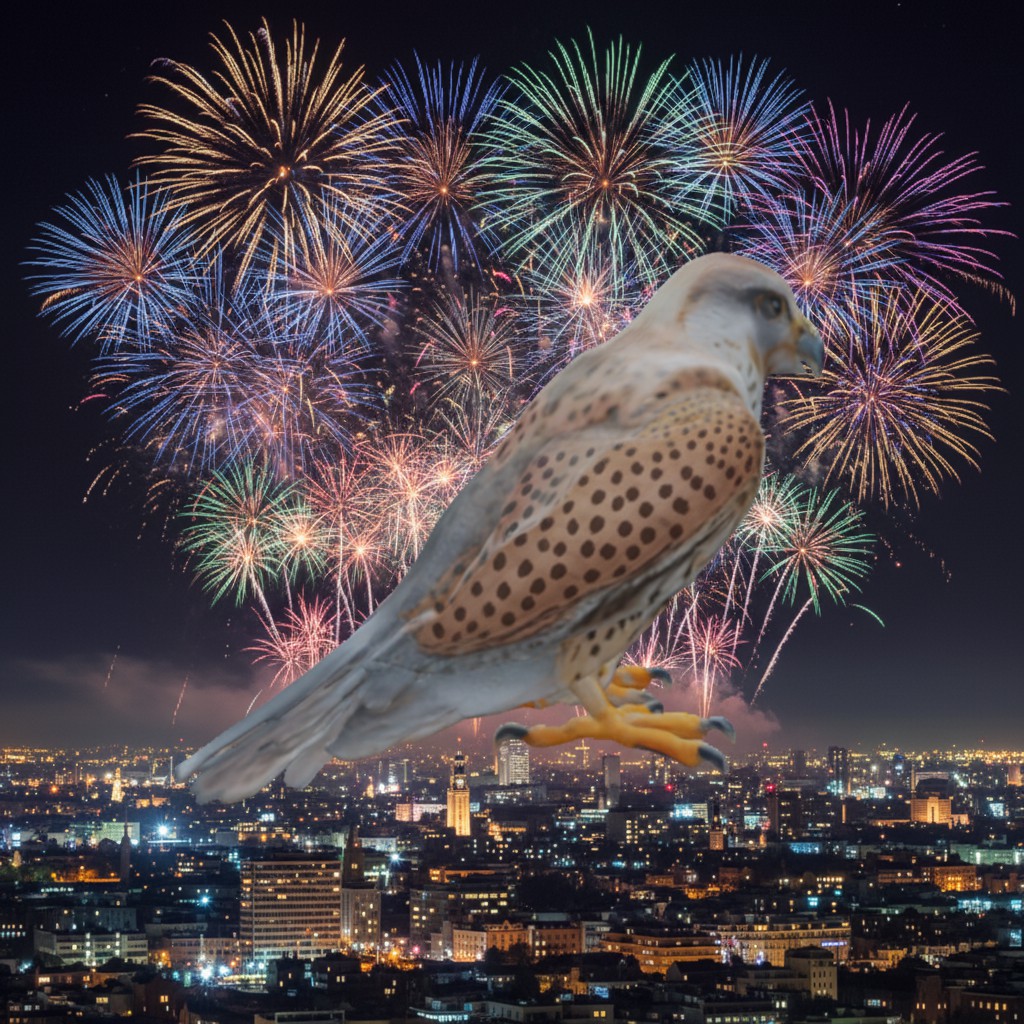} \\
\scriptsize (a) Incorrect scale &
\scriptsize (b) Viewpoint mismatch &
\scriptsize (c) Object floating \\
\end{tabular}
\caption{\textbf{Excluded augmented images.} (a) Object scale incorrect for scene depth, (b) background viewpoint mismatch, (c) object floating without grounding.}
\label{fig:dataset_bad_examples}
\end{figure}

We generate 3-5 detailed object descriptions per asset and curate 30 background prompts categorized by context (on land, on water, in air). During augmentation, we randomly sample object-background combinations to produce diverse, realistic appearances with varied environments. We filter the results to remove images with incorrect object pose, prompt misalignment (e.g., object scale incorrect for scene depth, background angle mismatched with object viewpoint), or physical implausibilities (e.g., floating objects), yielding 6,559 high-quality augmented images. \Cref{fig:dataset_bad_examples} shows examples of failures. 
\Cref{fig:dataset_examples} shows more examples of rendered training images and training images augmented to include backgrounds. \Cref{tab:dataset_prompts} shows the captions for the images.

\noindent \textbf{Captions.}
For the rendered images, we use the captions generated for the 3{,}111 objects as the text prompt, appended with a viewpoint token. For the photorealistic augmented images, we use the combination of the detailed object description and the background augmentation prompts as the text prompt, appended with a viewpoint token. The detailed descriptions and the background augmentation prompts are the same as the ones used for the Gemini 2.5 Flash Image when creating the augmented image.


\begin{table*}[h]
\centering
\small
\caption{\textbf{Viewpoint regressor accuracy} across the test split of four datasets. We report mean and median angular errors (degrees) and radius error (normalized).}
\label{tab:pose_regressor_accuracy}
\setlength{\tabcolsep}{5.5pt}
\begin{tabular}{lcccccccccc}
\toprule
\multirow{2}{*}{Dataset} &
\multicolumn{2}{c}{Azimuth$\downarrow$} &
\multicolumn{2}{c}{Elevation$\downarrow$} &
\multicolumn{2}{c}{Radius$\downarrow$} &
\multicolumn{2}{c}{Yaw$\downarrow$} &
\multicolumn{2}{c}{Pitch$\downarrow$} \\
\cmidrule(lr){2-3}
\cmidrule(lr){4-5}
\cmidrule(lr){6-7}
\cmidrule(lr){8-9}
\cmidrule(lr){10-11}
 & Mean & Median & Mean & Median & Mean & Median & Mean & Median & Mean & Median \\
\midrule
ControlNetPlus Augmentation   & 4.16 & 2.12 & 2.99 & 2.17 & 0.043 & 0.031 & 0.498 & 0.380 & 0.512 & 0.390 \\
Rendered Dataset              & 2.65 & 1.65 & 2.25 & 1.64 & 0.031 & 0.022 & 0.434 & 0.361 & 0.430 & 0.350 \\
Photorealistic Dataset        & 11.14 & 4.41 & 5.40 & 4.24 & 0.084 & 0.061 & 0.911 & 0.635 & 0.958 & 0.684 \\
Compass Control Training Set  & 9.54 & 5.14 & -- & -- & -- & -- & -- & -- & -- & -- \\
\bottomrule
\end{tabular}
\end{table*}

\section{Testing Dataset}
For evaluation of viewpoint accuracy and CLIP similarity, we use the 11 test objects (easy set) from Compass Control~\cite{Parihar_2025_CVPR} and introduce 26 additional objects ({\bf diverse set}) spanning broader categories: common animals (dog, cat, horse, cow, rabbit), rare animals (okapi, red panda, shoebill), vehicles (car, motorcycle, fighter jet, helicopter, buggy, snowmobile, gundam), furniture (chair), people (girl, woman, boy, man, elderly, Santa Claus, skeleton), and mythical creatures (phoenix, unicorn, mermaid). Notably, 11 of the additional objects (okapi, red panda, shoebill, buggy, snowmobile, gundam, Santa Claus, skeleton, phoenix, unicorn, mermaid) do not appear in our training data, testing generalization to unseen categories. For the diverse set (26 objects), we generate three descriptive phrases to test fine-grained prompt following. By combining 37 objects and background prompts, we have 555 unique prompt-object pairs. For each combination, we sample 10 random viewpoints, totaling 5,550 test samples.

\section{More Examples of Overfitting by Compass Control}\label{sup:compass}
\Cref{fig:overfitting_compass} shows more examples of Compass Control overfitting to its training objects. For example, when it is asked to generate Santa Claus, it generates a shoe with Santa Claus appearance. \Cref{fig:compass_pie} further illustrates the distribution of these overfitting cases. Specifically, we examine Compass Control’s outputs for the Santa Claus, rabbit, and dolphin prompts in our test set and identify the mismatches—cases where the generated object does not match the prompt. In these failures, Compass Control often produces shapes resembling objects from its training set (lions, ostriches, teddy bears, shoes, and sofas), instead of the object named in the test prompt.
As a comparison, our results follow the prompts for both categories included in our training set (e.g., rabbit and dolphin) and novel categories not included in our training set (e.g., Santa Claus).

\section{Viewpoint Regressor}
\label{sup:regressor}
The regressor we use in evaluation is built on a pretrained ResNet-34~\cite{he2016deep} backbone appended with three linear layers with ReLU~\cite{agarap2018deep} activation. The regressor outputs a 6-dimensional vector representing the viewpoint: $[\sin(\theta_{\text{az}}), \cos(\theta_{\text{az}}), \theta_{\text{el}}, r, \theta_{\text{pitch}}, \theta_{\text{yaw}}] \in \mathbb{R}^{6}$. We normalize the $[\sin(\theta_{\text{az}}), \cos(\theta_{\text{az}})]$ component to have a norm of 1. We train the network to estimate the pose of objects using a range of data with known poses generated by (i) ControlNetPlus~\cite{controlnetplus} provided Canny edge maps of the rendered 37 testing objects (ii) rendered dataset (iii) photorealistic augmented dataset, and (iv) Compass Control training dataset. For the Compass Control training dataset, we only have the annotation for the $
\theta_{\text{az}}$; therefore, we only backpropagate loss on the $[\sin(\theta_{\text{az}}), \cos(\theta_{\text{az}})]$ output. We hold out 10\% of each subset for validation and measure azimuth estimation errors of $4.16^\circ$ on images of type (i), $2.64^\circ$ for type (ii), $11.14^\circ$ for type (iii), and $9.53^\circ$ for type (iv). See \cref{tab:pose_regressor_accuracy} for more details.


\section{More Qualitative Examples}
\begin{figure*}
\centering
\resizebox{\textwidth}{!}{%
\setlength{\tabcolsep}{2pt}
\begin{tabular}{c|cccccccc}
 & 10° & 55° & 100° & 145° & 190° & 235° & 280° & 325° \\
0° &
\includegraphics[width=0.12\textwidth]{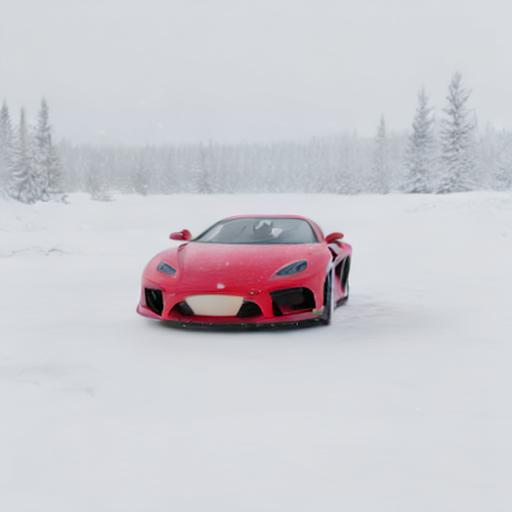} &
\includegraphics[width=0.12\textwidth]{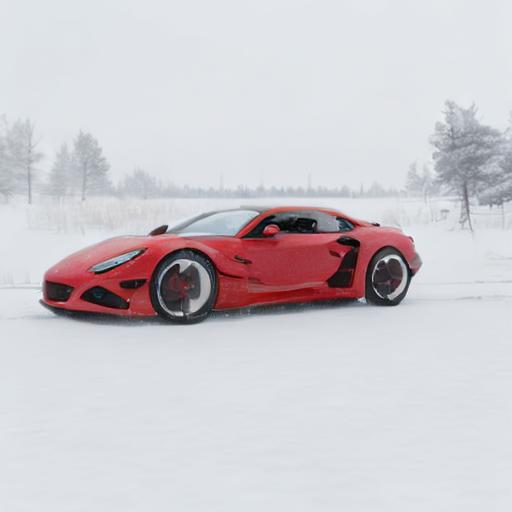} &
\includegraphics[width=0.12\textwidth]{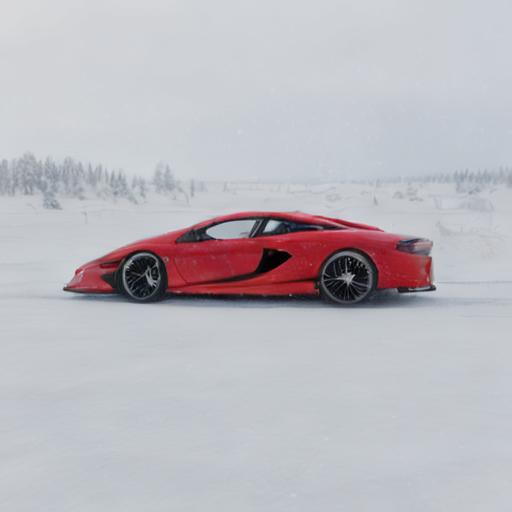} &
\includegraphics[width=0.12\textwidth]{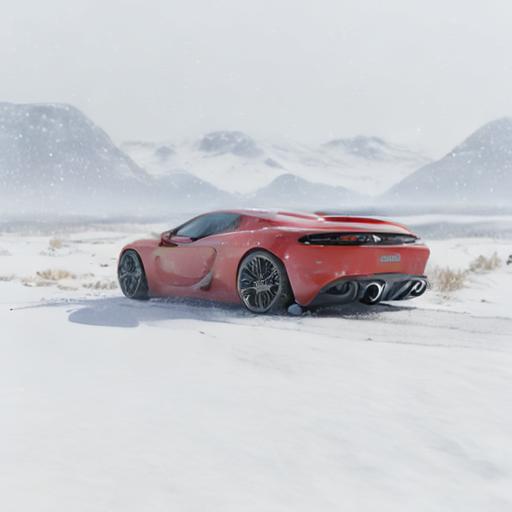} &
\includegraphics[width=0.12\textwidth]{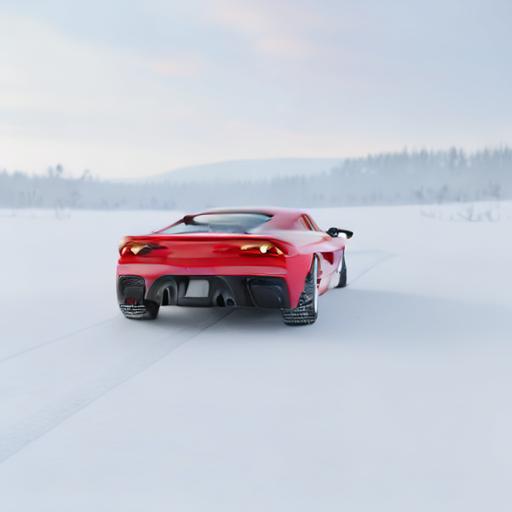} &
\includegraphics[width=0.12\textwidth]{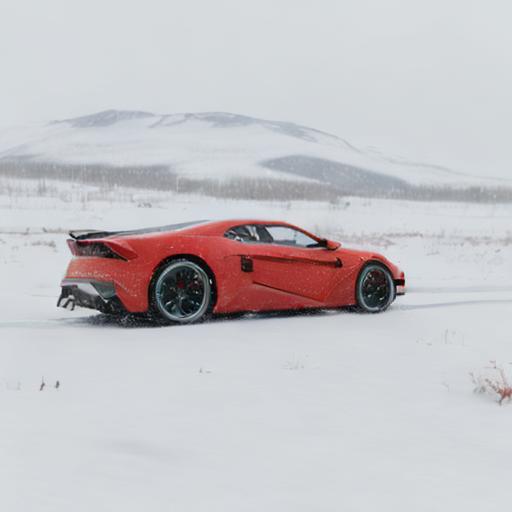} &
\includegraphics[width=0.12\textwidth]{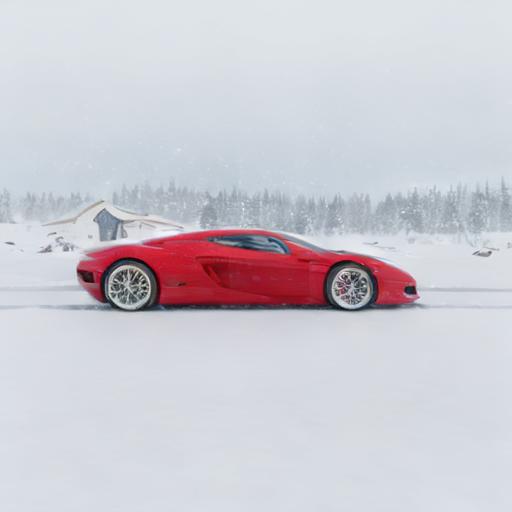} &
\includegraphics[width=0.12\textwidth]{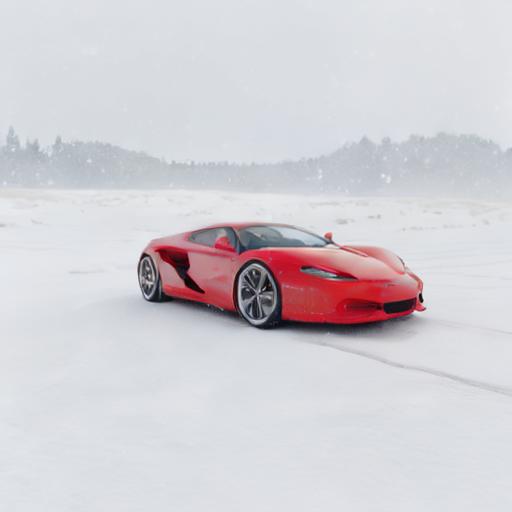} \\
15° &
\includegraphics[width=0.12\textwidth]{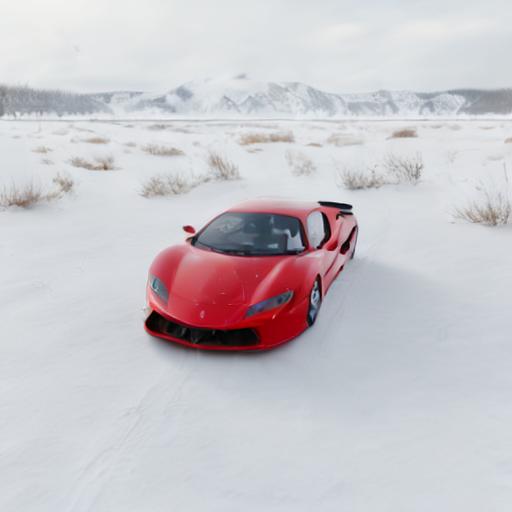} &
\includegraphics[width=0.12\textwidth]{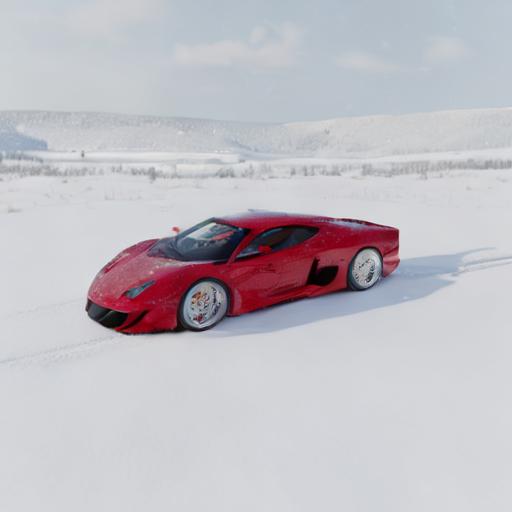} &
\includegraphics[width=0.12\textwidth]{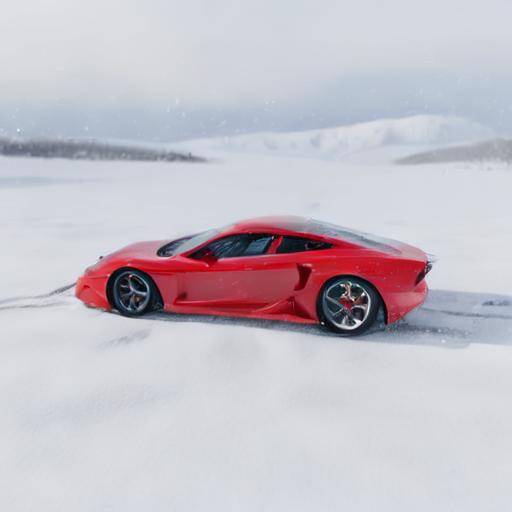} &
\includegraphics[width=0.12\textwidth]{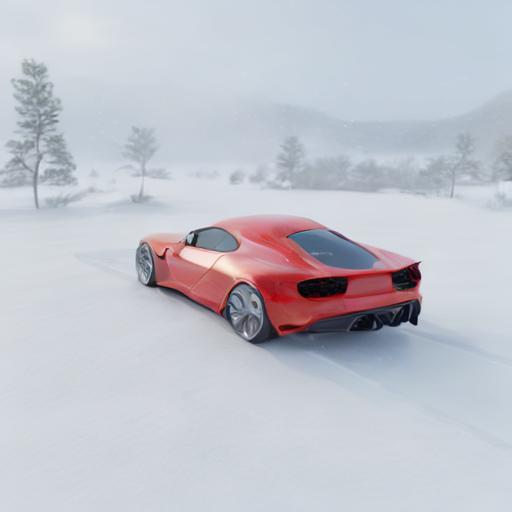} &
\includegraphics[width=0.12\textwidth]{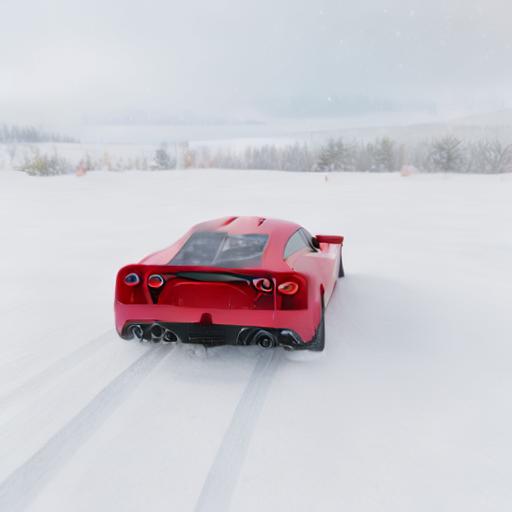} &
\includegraphics[width=0.12\textwidth]{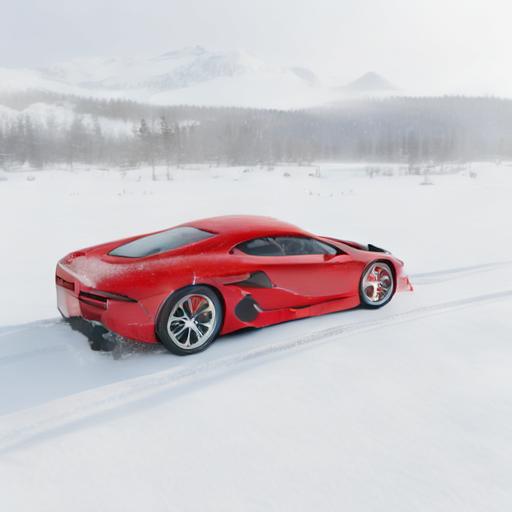} &
\includegraphics[width=0.12\textwidth]{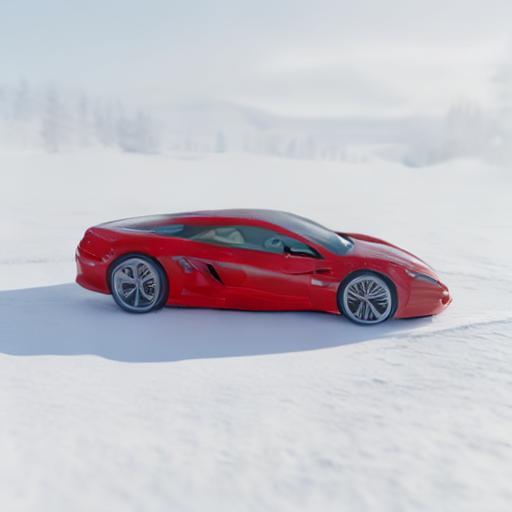} &
\includegraphics[width=0.12\textwidth]{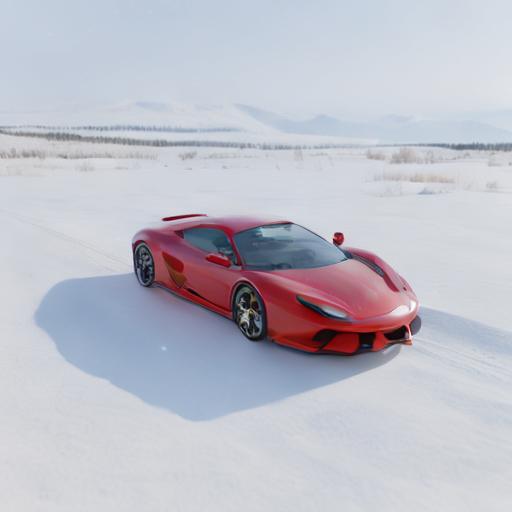} \\
30° &
\includegraphics[width=0.12\textwidth]{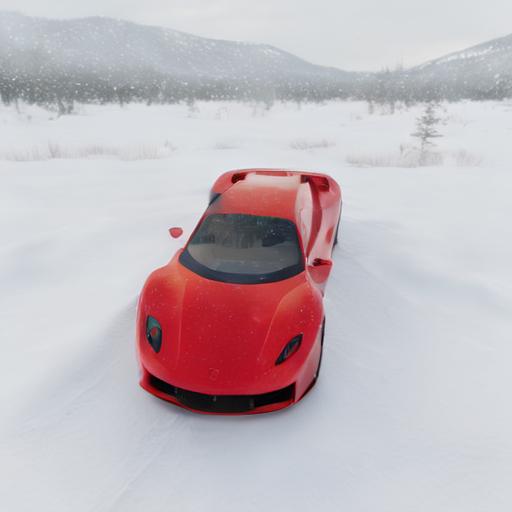} &
\includegraphics[width=0.12\textwidth]{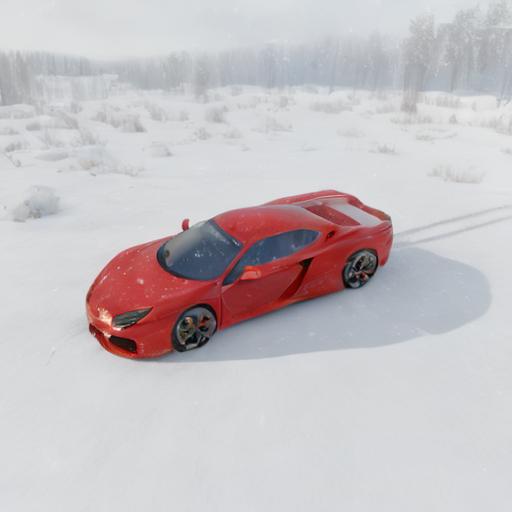} &
\includegraphics[width=0.12\textwidth]{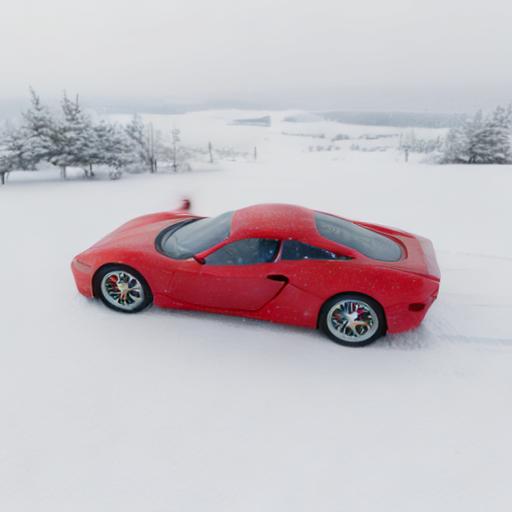} &
\includegraphics[width=0.12\textwidth]{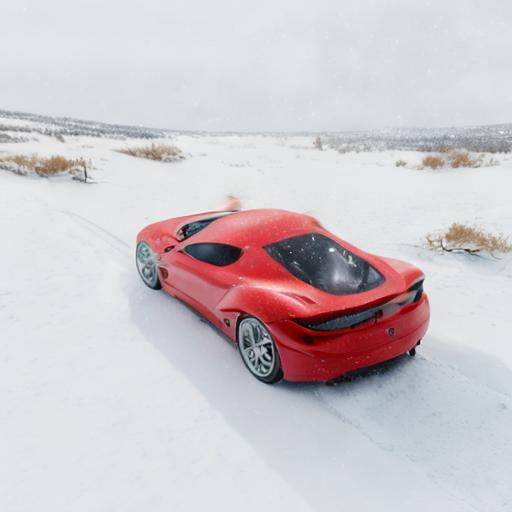} &
\includegraphics[width=0.12\textwidth]{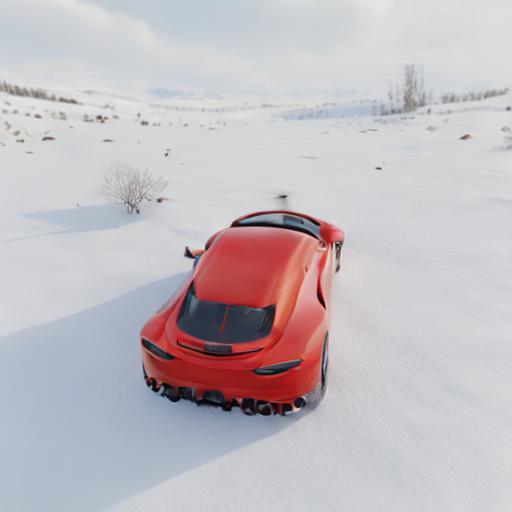} &
\includegraphics[width=0.12\textwidth]{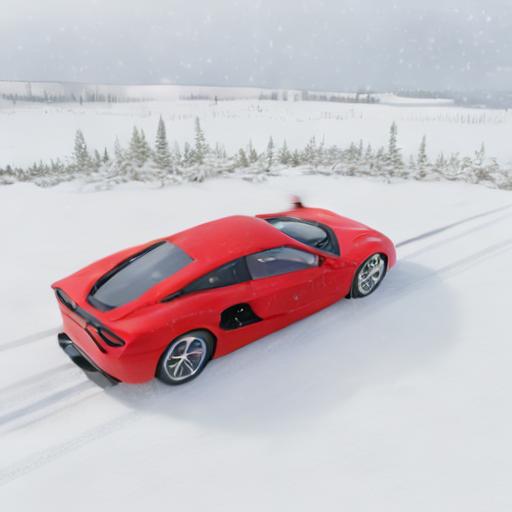} &
\includegraphics[width=0.12\textwidth]{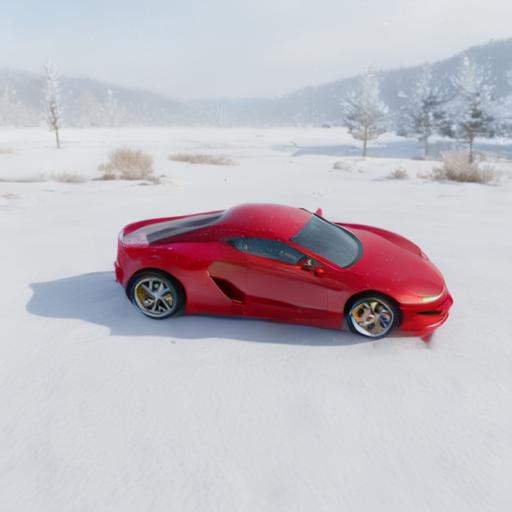} &
\includegraphics[width=0.12\textwidth]{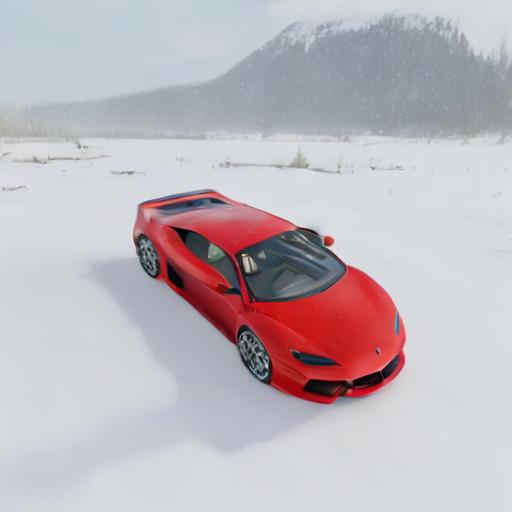} \\
45° &
\includegraphics[width=0.12\textwidth]{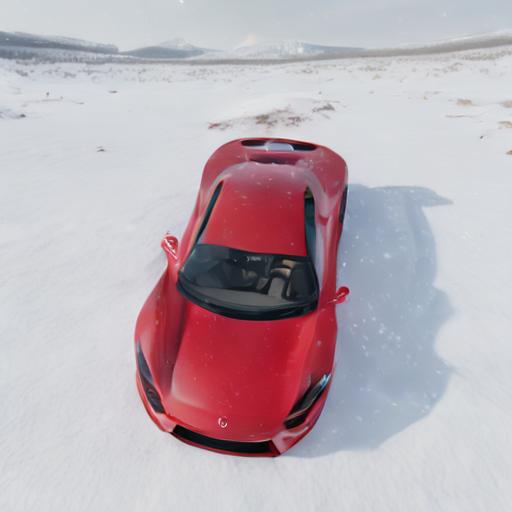} &
\includegraphics[width=0.12\textwidth]{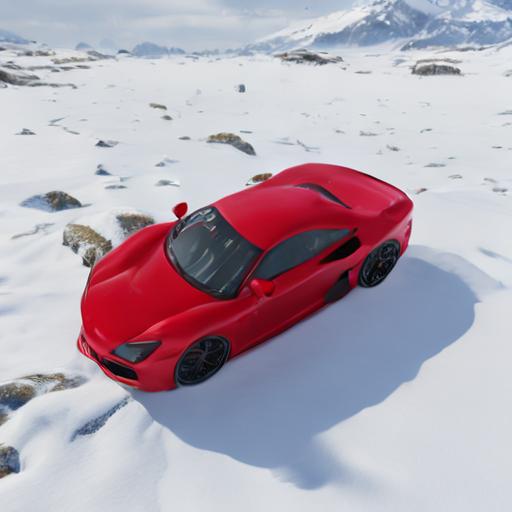} &
\includegraphics[width=0.12\textwidth]{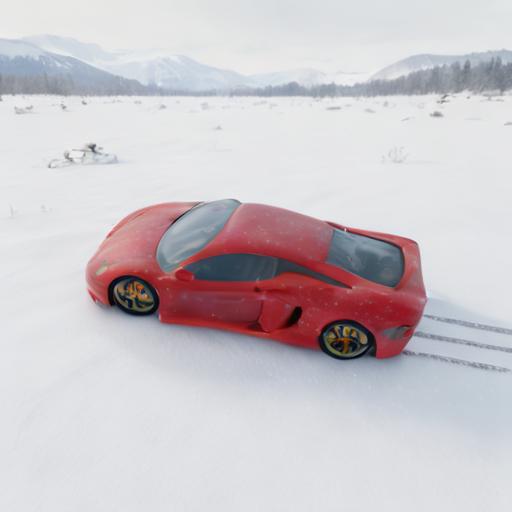} &
\includegraphics[width=0.12\textwidth]{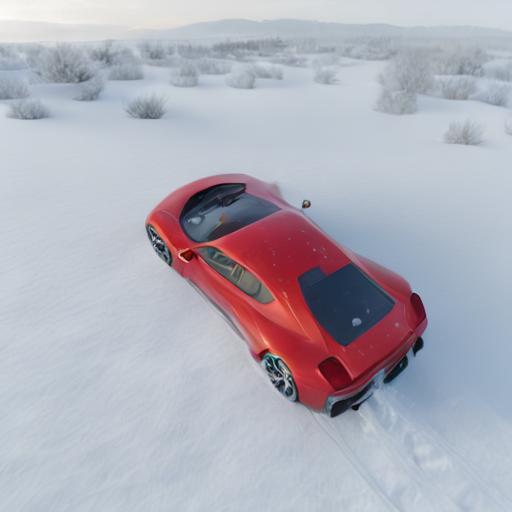} &
\includegraphics[width=0.12\textwidth]{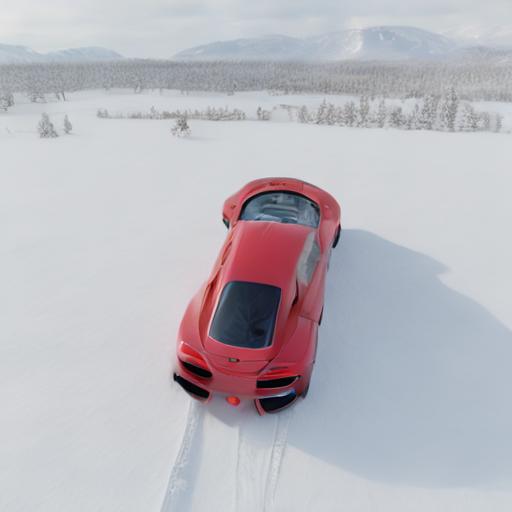} &
\includegraphics[width=0.12\textwidth]{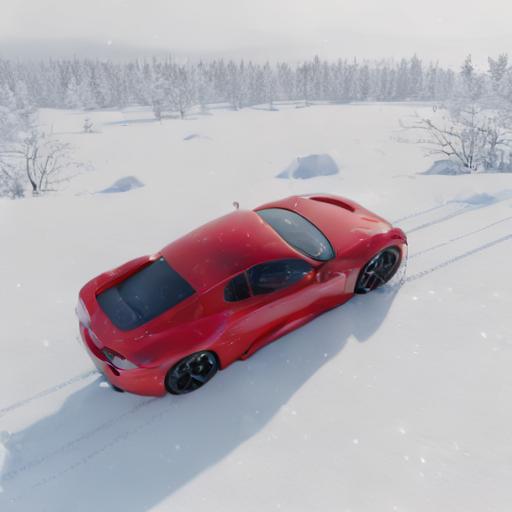} &
\includegraphics[width=0.12\textwidth]{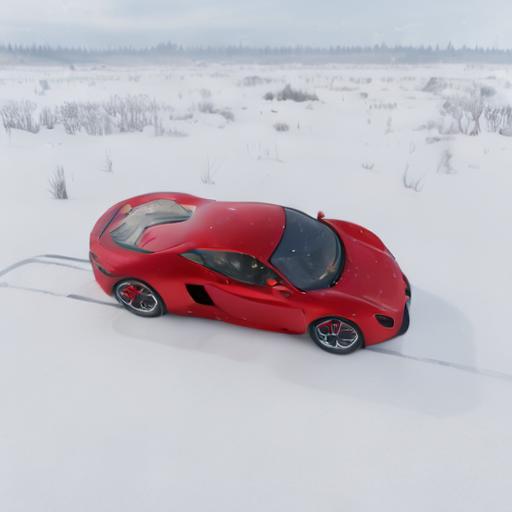} &
\includegraphics[width=0.12\textwidth]{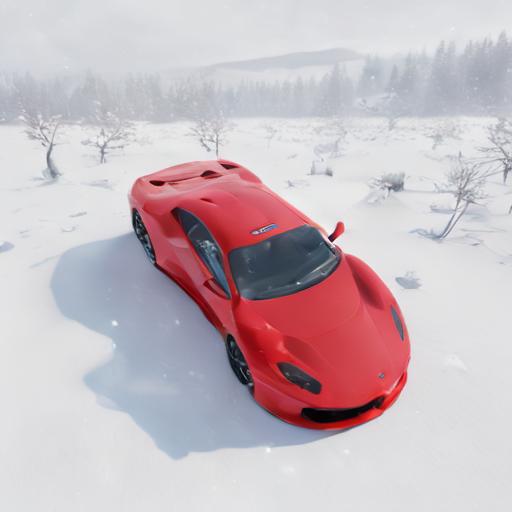} \\
\end{tabular}
}
\caption{\textbf{Generation results across azimuth and elevation.} Columns represent azimuth angles (10° to 325°) and rows represent elevation angles (0° to 45°). All examples use a fixed camera radius of 1.5 with pitch = 0° and yaw = 0°. All examples use the same seed 42.}
\label{fig:multiviewpoint}
\end{figure*}

\begin{figure*}[h]
\centering
\resizebox{0.7\textwidth}{!}{%
\setlength{\tabcolsep}{2pt}
\begin{tabular}{c|ccc|c}
 & -10° & 0° & 10° & 10° (r=2) \\
10° &
\includegraphics[width=0.18\textwidth]{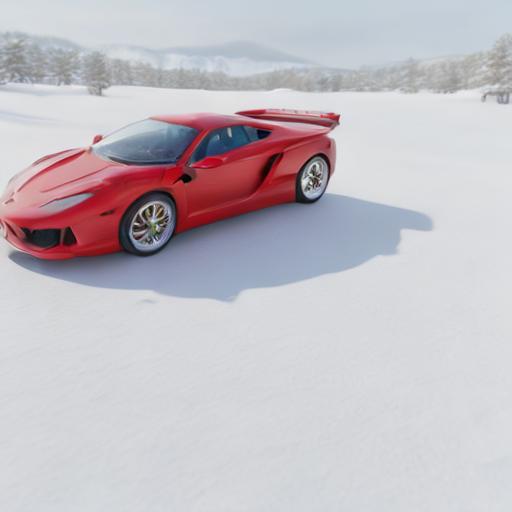} &
\includegraphics[width=0.18\textwidth]{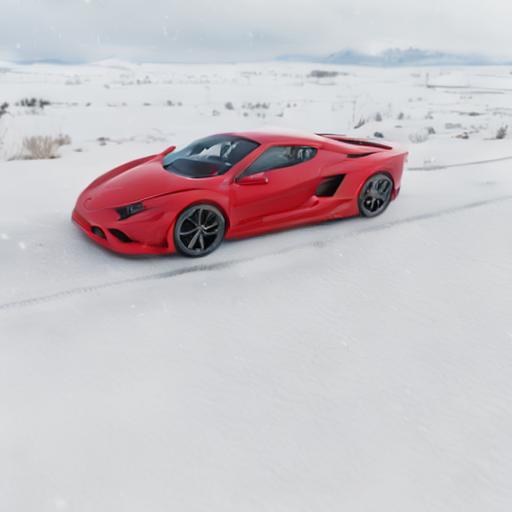} &
\includegraphics[width=0.18\textwidth]{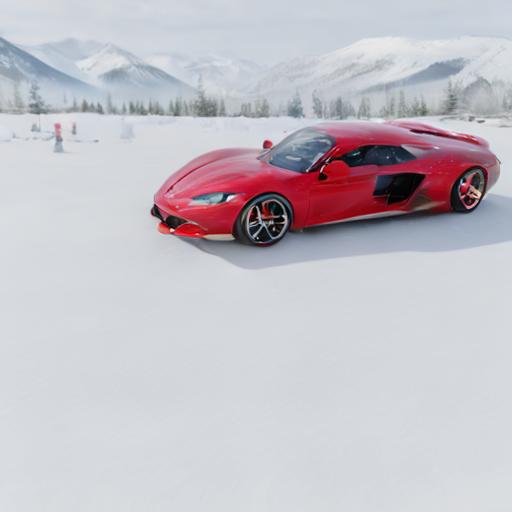} &
\includegraphics[width=0.18\textwidth]{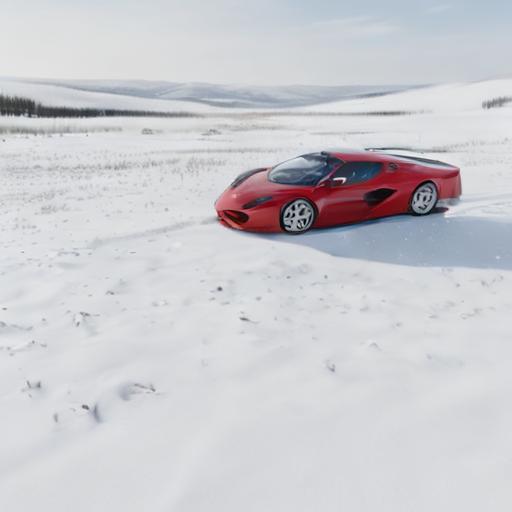} \\
0° &
\includegraphics[width=0.18\textwidth]{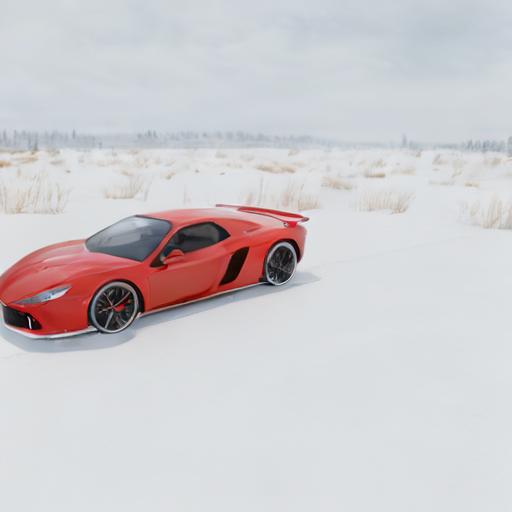} &
\includegraphics[width=0.18\textwidth]{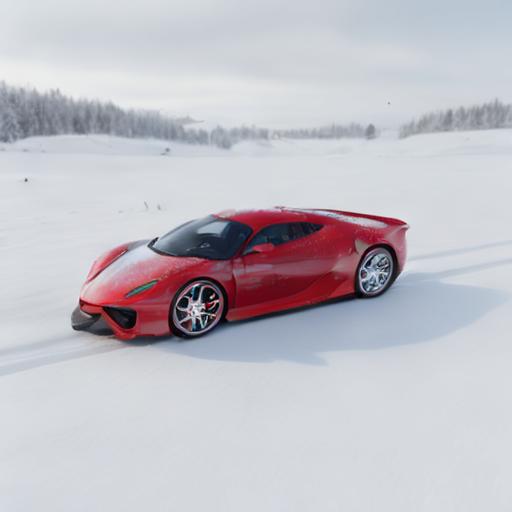} &
\includegraphics[width=0.18\textwidth]{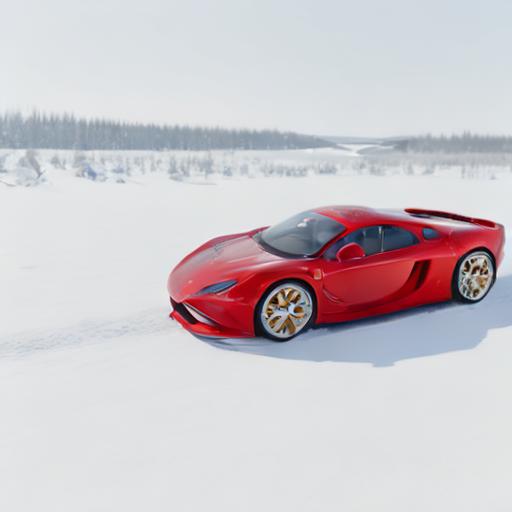} &
\includegraphics[width=0.18\textwidth]{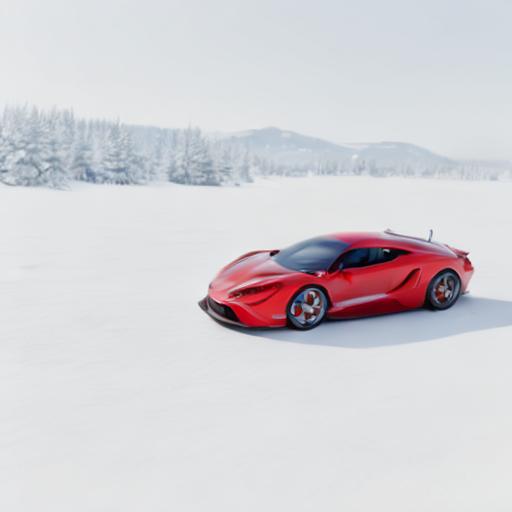} \\
-10° &
\includegraphics[width=0.18\textwidth]{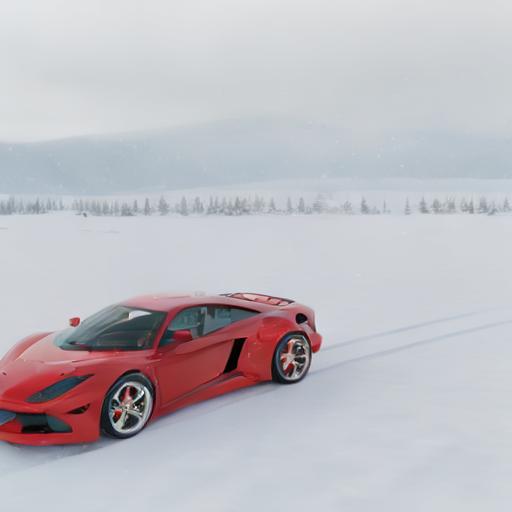} &
\includegraphics[width=0.18\textwidth]{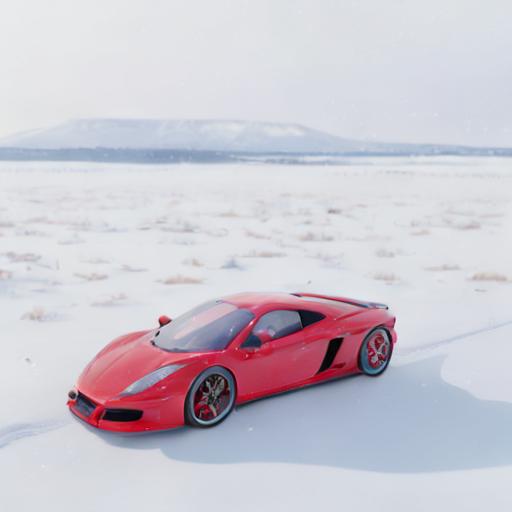} &
\includegraphics[width=0.18\textwidth]{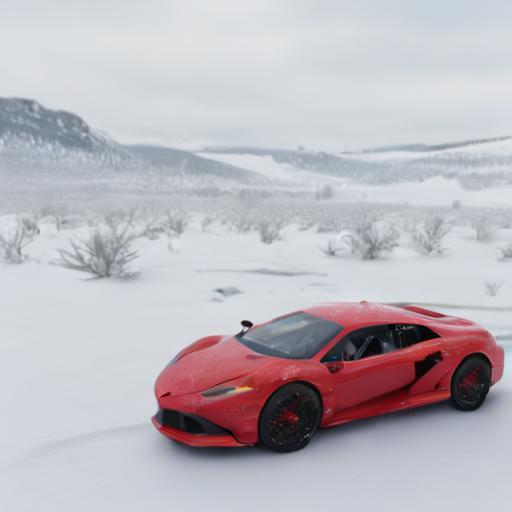} &
\includegraphics[width=0.18\textwidth]{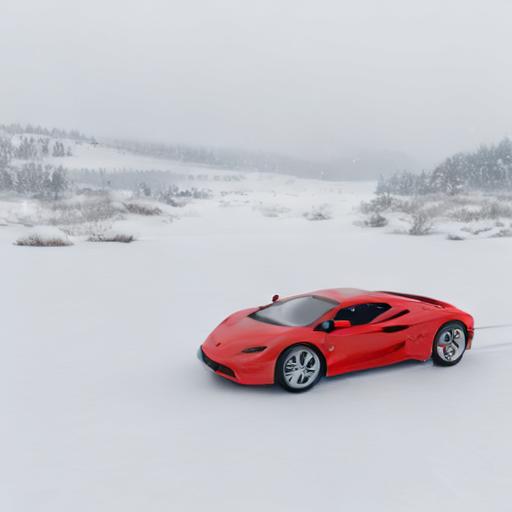} \\
\hline
-10° (r=2) &
\includegraphics[width=0.18\textwidth]{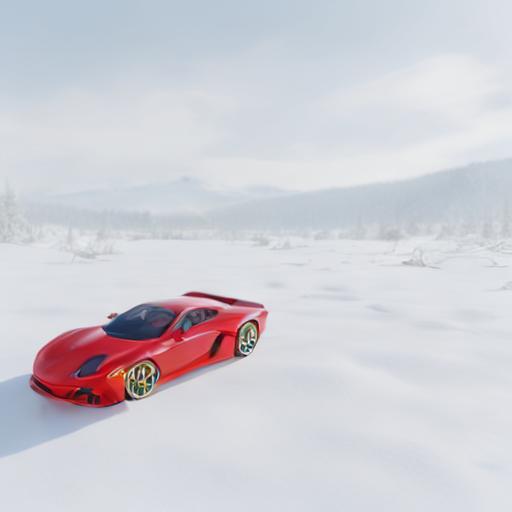} &
\includegraphics[width=0.18\textwidth]{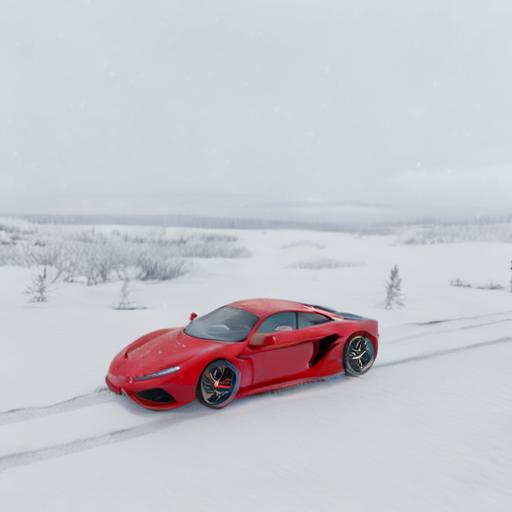} &
\includegraphics[width=0.18\textwidth]{qualitative_results/multi_pitch_yaw_radius_6/az+055.00_el+15.00_r+6.00_p-10.00_y+10.00.jpg} &
 \\
\end{tabular}
}
\caption{\textbf{Generation results across pitch, yaw, and radius.} Main 3×3 grid shows radius=1.5, extra row and column show radius=2.0. All examples use a fixed azimuth = 55° and elevation = 15°. All examples use the same seed 42.}
\label{fig:pitch_yaw}
\end{figure*}

\Cref{fig:multiviewpoint,fig:pitch_yaw} show examples of different camera parameters with the prompt ``A photo of a red sports car in a national reserve in a snowy landscape''.
\Cref{fig:more_qualitative_grid,fig:more_qualitative_grid_continued} present additional qualitative results on object categories not included in our training data. \Cref{fig:more_qualitative_grid_3} provides further examples for categories that are part of our training set.

\begin{figure*}[t]
\centering
\setlength{\tabcolsep}{2pt}
\renewcommand{\arraystretch}{1.0}

\begin{tabular}{@{}ccccc@{}}
\includegraphics[width=0.19\textwidth]{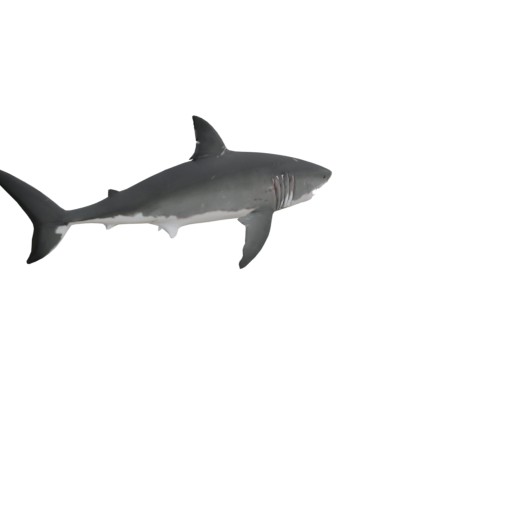} &
\includegraphics[width=0.19\textwidth]{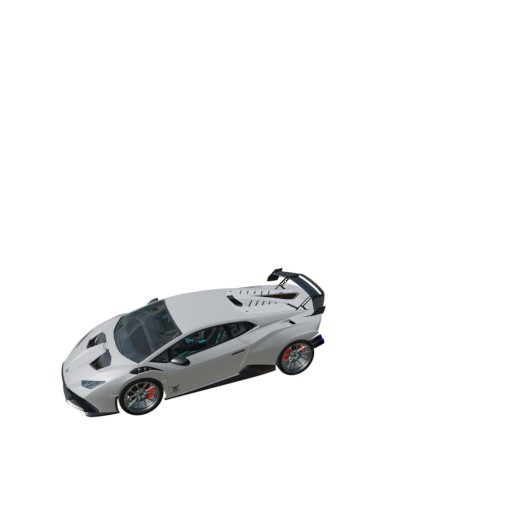} &
\includegraphics[width=0.19\textwidth]{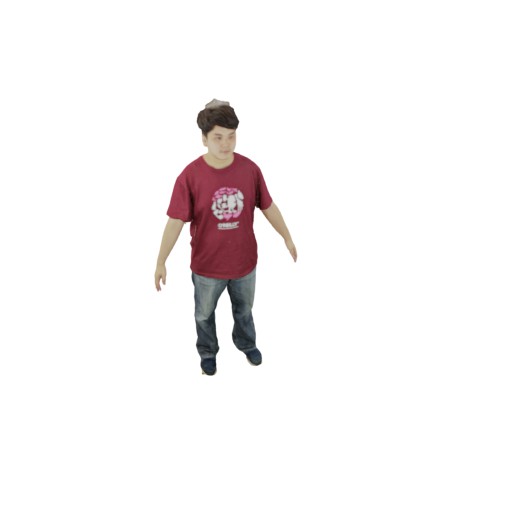} &
\includegraphics[width=0.19\textwidth]{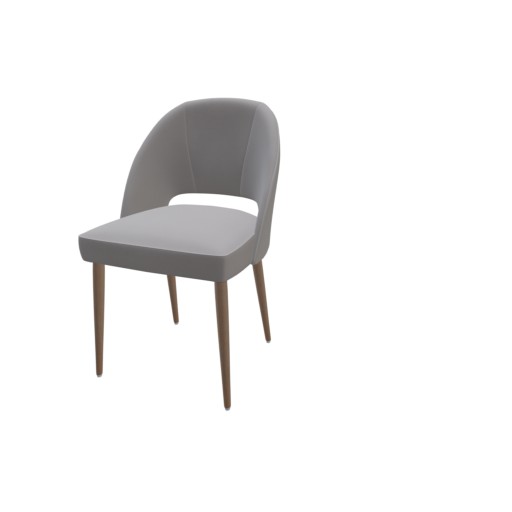} &
\includegraphics[width=0.19\textwidth]{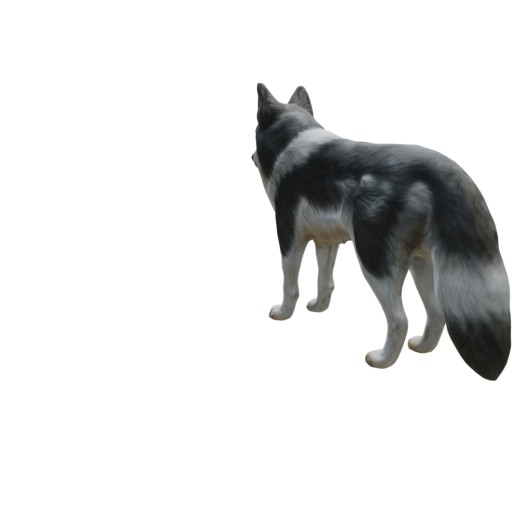} \\
(a) & (b) & (c) & (d) & (e) \\[4pt]

\includegraphics[width=0.19\textwidth]{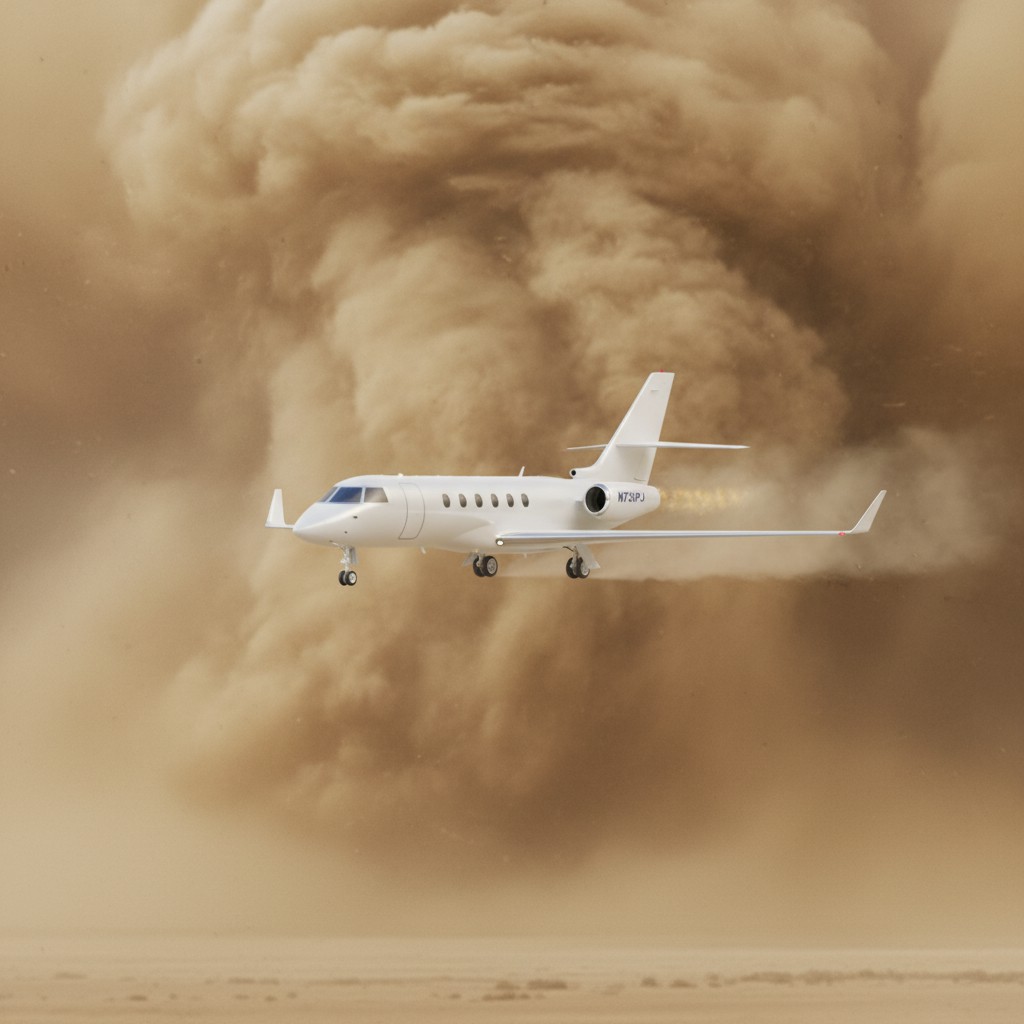} &
\includegraphics[width=0.19\textwidth]{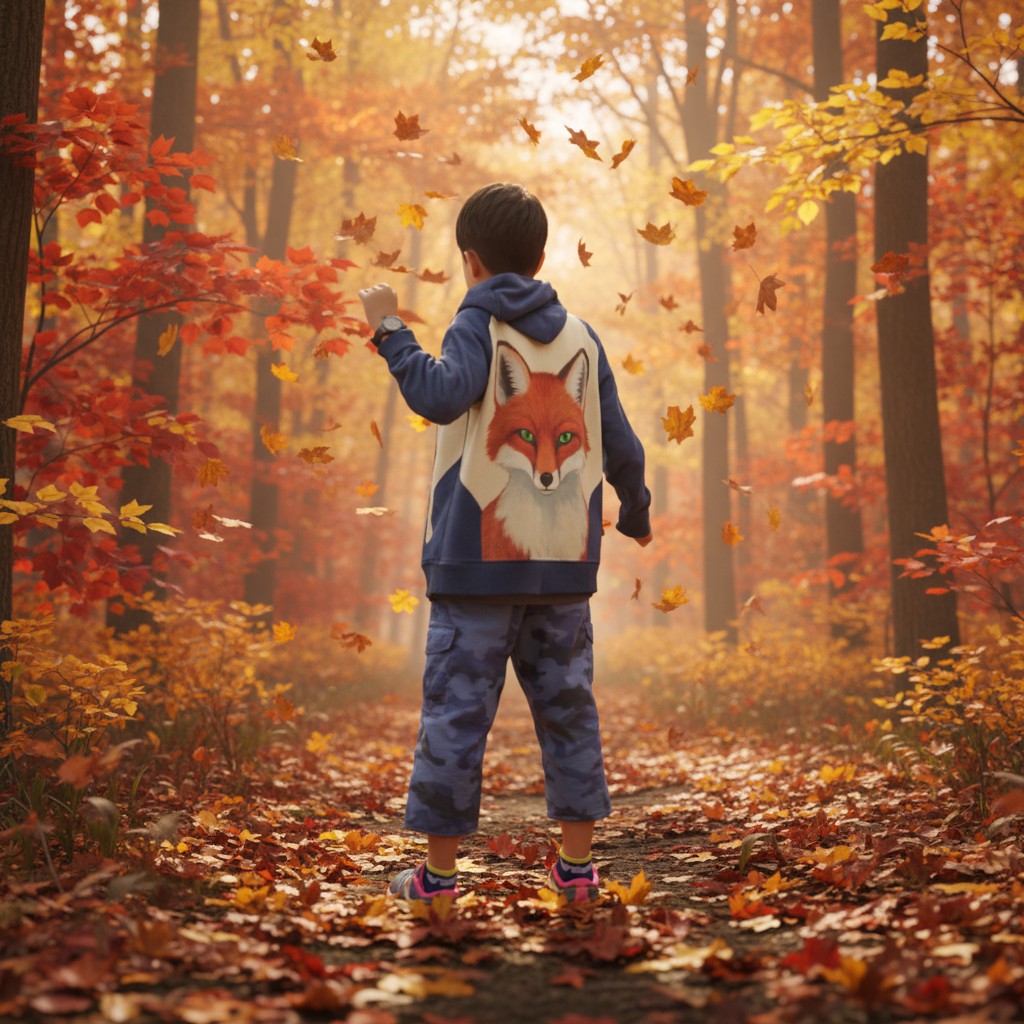} &
\includegraphics[width=0.19\textwidth]{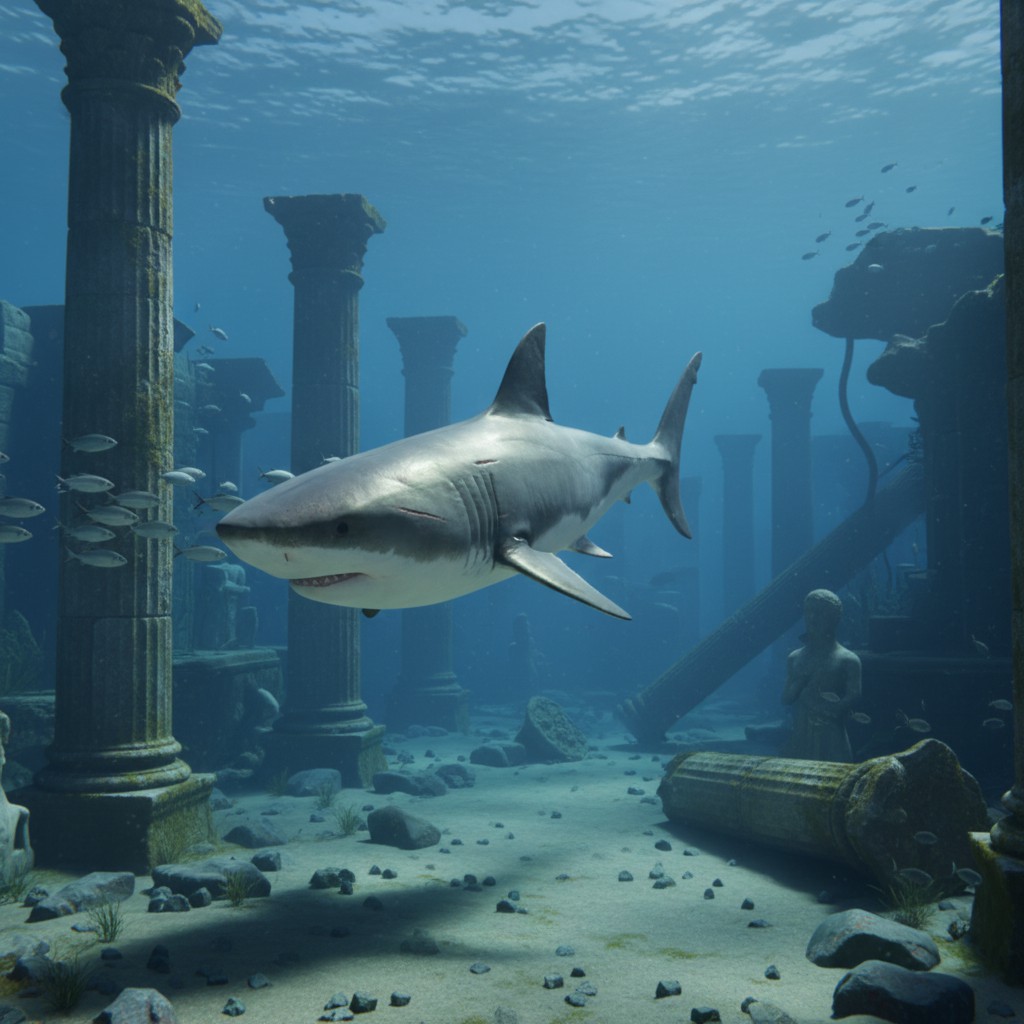} &
\includegraphics[width=0.19\textwidth]{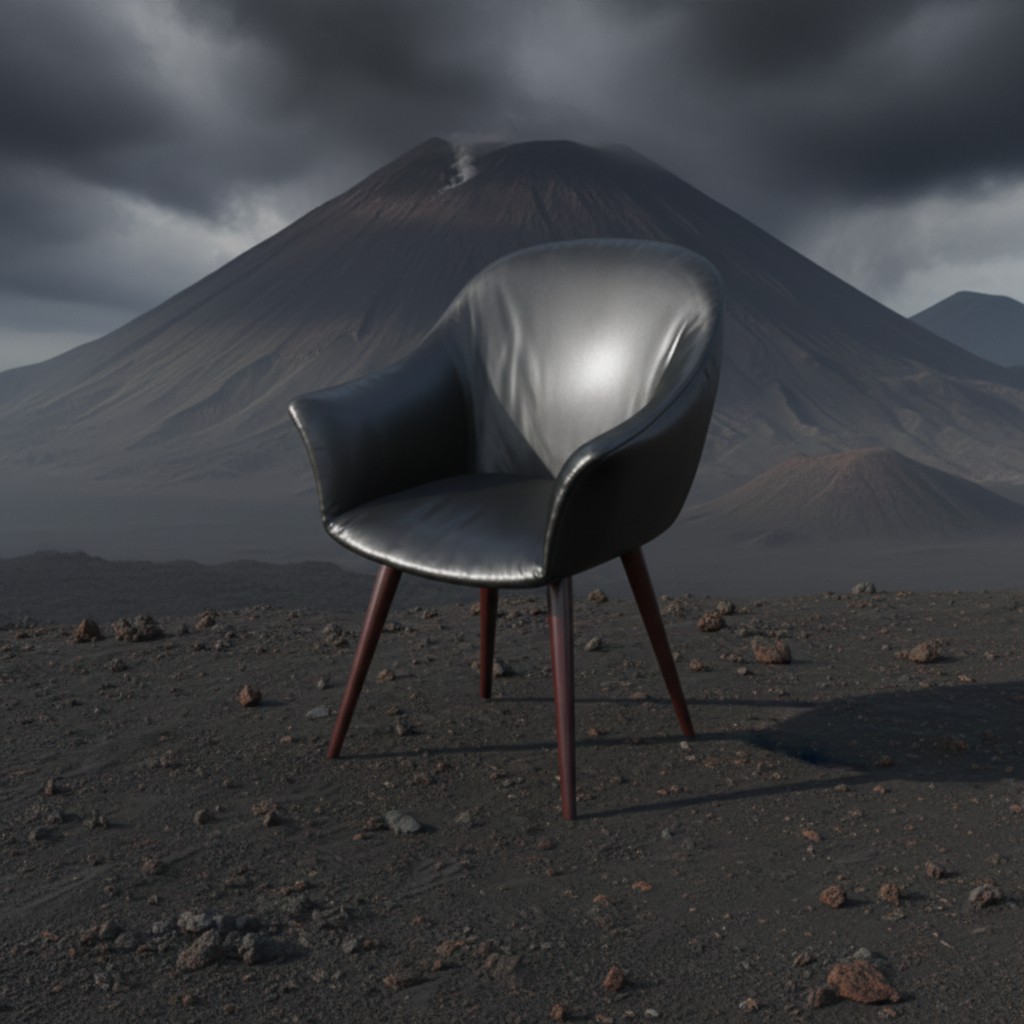} &
\includegraphics[width=0.19\textwidth]{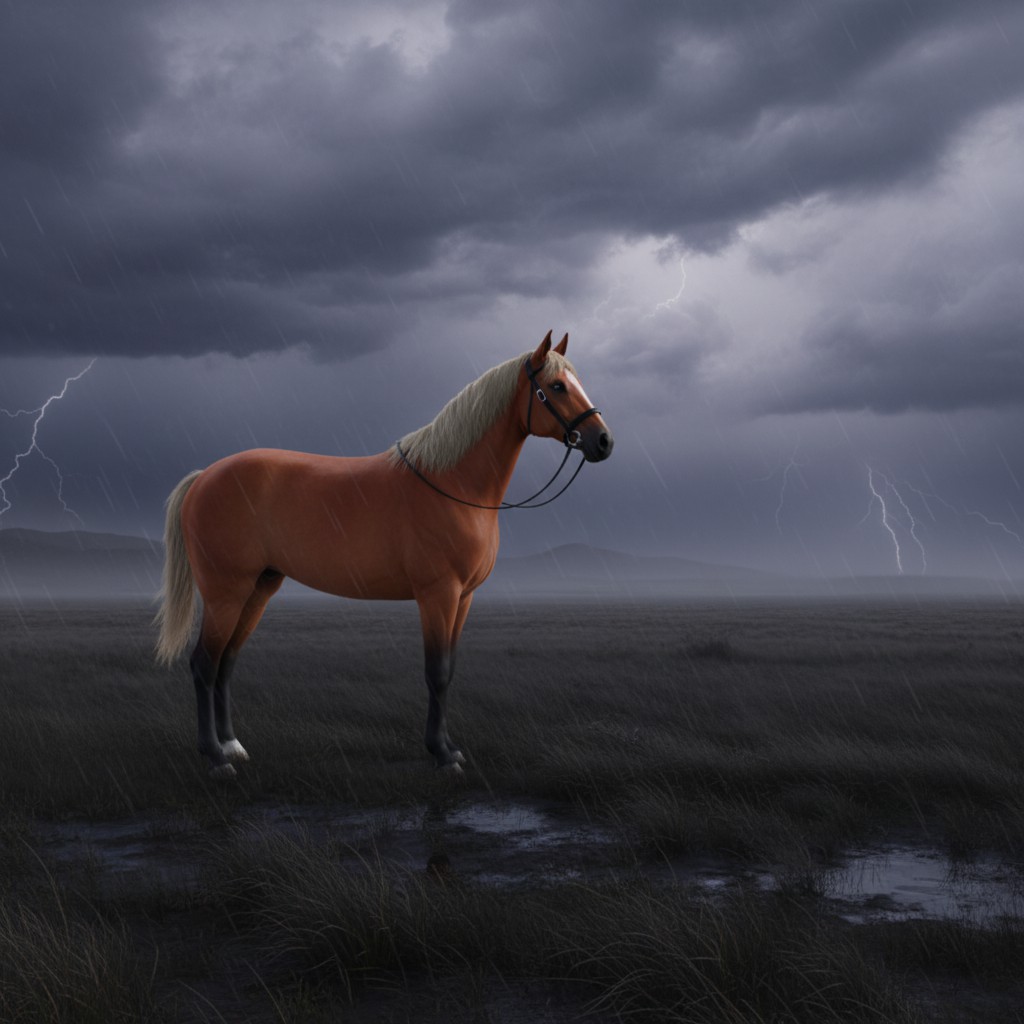} \\
(f) & (g) & (h) & (i) & (j)
\end{tabular}

\caption{\textbf{Training dataset examples.} Top row: rendered dataset. Bottom row: photorealistic augmented dataset.}
\label{fig:dataset_examples}
\end{figure*}

\begin{table*}[t]
\centering
\small
\caption{\textbf{Text prompts} for images in \Cref{fig:dataset_examples}, listed left-to-right, top-to-bottom.}
\label{tab:dataset_prompts}
\begin{tabular}{cp{0.9\textwidth}}
\toprule
ID & Prompt \\
\midrule
(a) & A 3D model of a great white shark with dark gray back. The shark has white underside, pointed snout, gill slits, and powerful tail fin. \\
(b) & A 3D rendered white sports car with large rear wing. The car features aggressive aerodynamics, air intakes, and track-focused modifications. \\
(c) & A boy in burgundy t-shirt with graphic and gray jeans. He has dark hair and wears dark sneakers, standing casually. \\
(d) & A modern dining chair with beige fabric and curved open backrest. The chair has wooden legs and a distinctive circular cutout in the backrest. \\
(e) & A 3D model of a husky with black and white fur. The dog has pointed ears and curled tail with blue eyes, standing. \\
(f) & Object: A business jet with private registration, rear-mounted engines, and standard wing tips. \newline Background: A sky filled with swirling dust during a distant sandstorm. \\
(g) & Object: A boy with brown skin and graphic print t-shirt underneath. \newline Background: An autumn forest with colorful, falling leaves. \\
(h) & Object: A shark with dorsal fin, cream underbelly, and robust body. \newline Background: A submerged ancient city ruin. \\
(i) & Object: An armchair with padded armrests, low profile back, and tapered wooden legs. \newline Background: A volcanic landscape with dark ash. \\
(j) & Object: A horse wearing a bridle with dark legs. \newline Background: A plain under a dramatic, stormy sky. \\
\bottomrule
\end{tabular}
\end{table*}

\newcommand{\imgcellnew}[1]{%
  \includegraphics[height=\qualheight]{qualitative_results/more_qualitative_results/#1}%
}

\newcommand{\qualrownew}[2]{
  \imgcellnew{#1_camera} &
  \imgcellnew{#1_3d} &
  \imgcellnew{#1_controlnet} &
  \imgcellnew{#1_novelview} &
  \imgcellnew{#1_compass} &
  \imgcellnew{#1_final20} \\[-6pt]
  \multicolumn{6}{l}{\scriptsize \emph{Prompt:} #2} \\[2pt]
}

\begin{figure*}[t]
\small
\centering
\setlength{\tabcolsep}{1.5pt}
\renewcommand{\arraystretch}{1.12}
\begin{tabular}{ZYYYYY}
\toprule
\textbf{Camera Spec} &
\textbf{3D Render \hspace{5cm} (GT View)} &
\textbf{ControlNet} &
\textbf{SV-Camera} &
\textbf{Compass Control} &
\textbf{\;Ours\;} \\
\midrule




\qualrownew{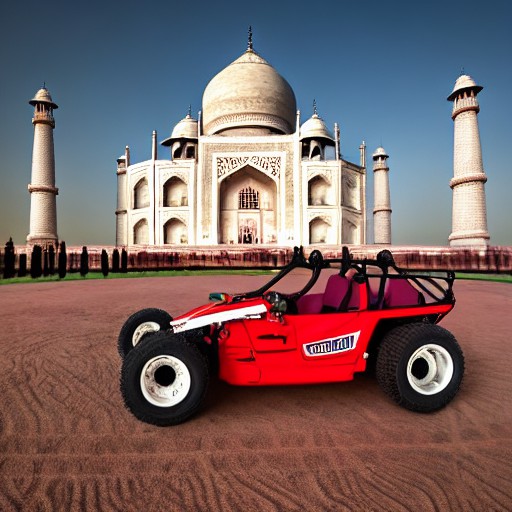}{
  A photo of red off-road buggy with large tires in front of the Taj Mahal
}

\qualrownew{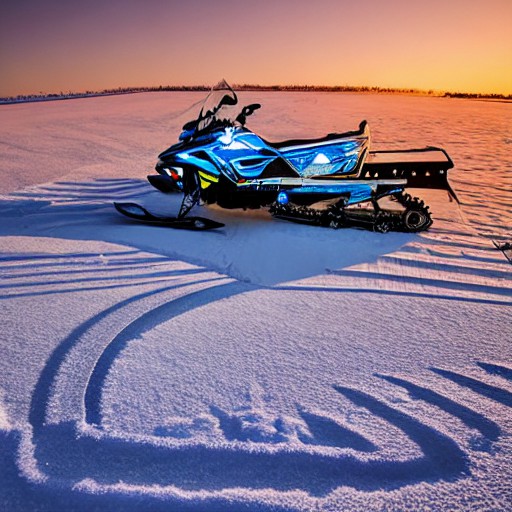}{
  A photo of blue snowmobile with track and headlight in front of the Taj Mahal
}

\qualrownew{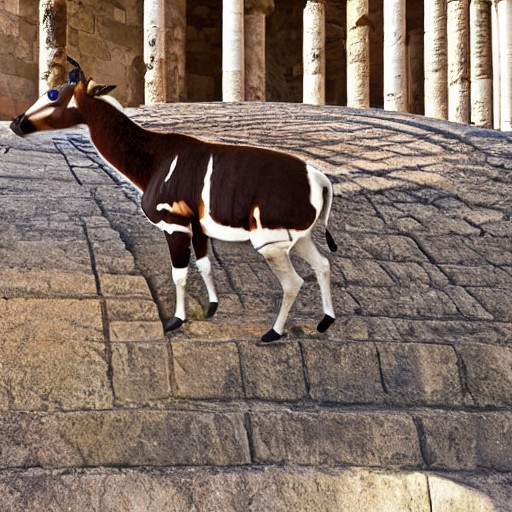}{
  A photo of okapi with velvety brown fur and bold stripes in an ancient Greek temple ruin, with broken columns and weathered stone steps
}

\qualrownew{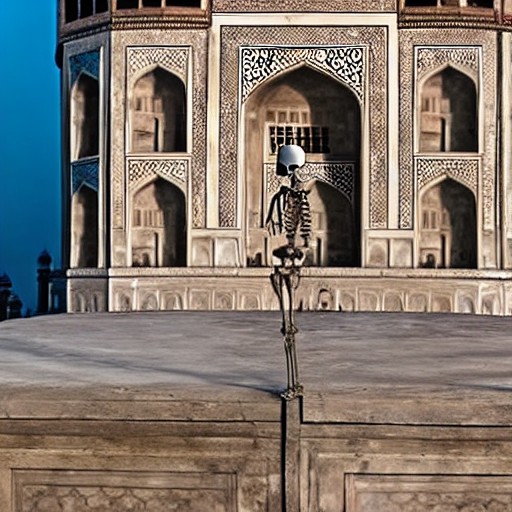}{
  A photo of human skeleton standing upright with arms at sides in front of the Taj Mahal
}

\qualrownew{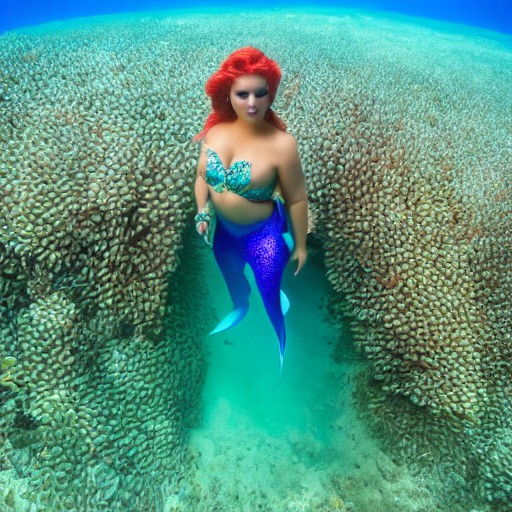}{
  A photo of mermaid with scales shimmering in green and blue in a vibrant coral reef, teeming with colorful tropical fish
}

\qualrownew{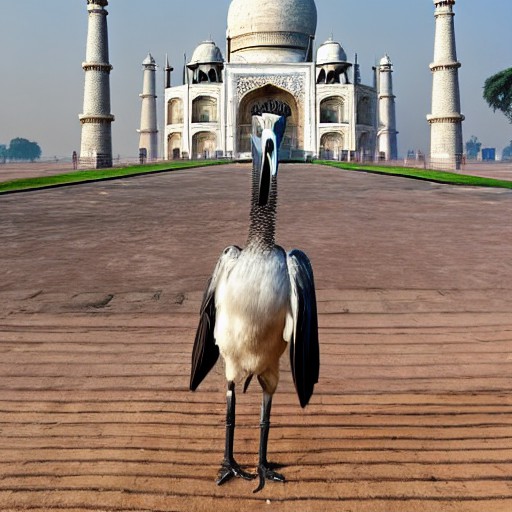}{
  A photo of shoebill with distinctive large bill and tall stance in front of the Taj Mahal
}

\bottomrule
\end{tabular}
\caption{\textbf{More qualitative comparisons (Part 1).} Each row pair shows images (top) and the corresponding prompt (bottom). The first two columns display the camera frustum (3D illustration) and a ground truth 3D rendering from a similar 3D object to the prompt. The remaining columns show results from different methods: ControlNet~\cite{zhang2023adding}, Stable-Virtual-Camera~\cite{zhou2025stable}, Compass Control~\cite{Parihar_2025_CVPR}, and our method. The object types in the prompts are not included in our training dataset.}
\label{fig:more_qualitative_grid}
\end{figure*}

\begin{figure*}[t]
\small
\centering
\setlength{\tabcolsep}{1.5pt}
\renewcommand{\arraystretch}{1.12}
\begin{tabular}{ZYYYYY}
\toprule
\textbf{Camera Spec} &
\textbf{3D Render \hspace{5cm} (GT View)} &
\textbf{ControlNet} &
\textbf{SV-Camera} &
\textbf{Compass Control} &
\textbf{\;Ours\;} \\
\midrule


\qualrownew{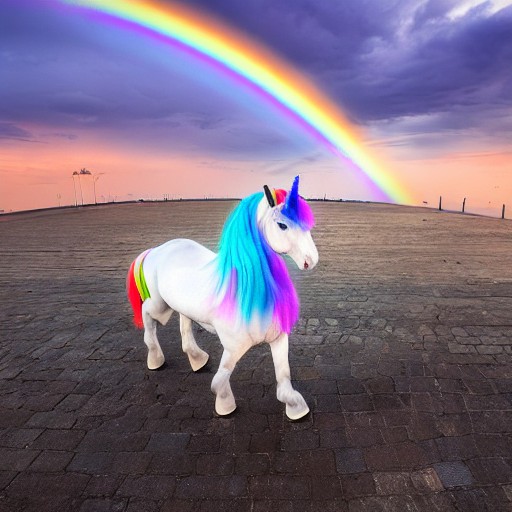}{
  A photo of graceful unicorn with rainbow mane and tail on the streets of Venice, with the sun setting in the background
}

\qualrownew{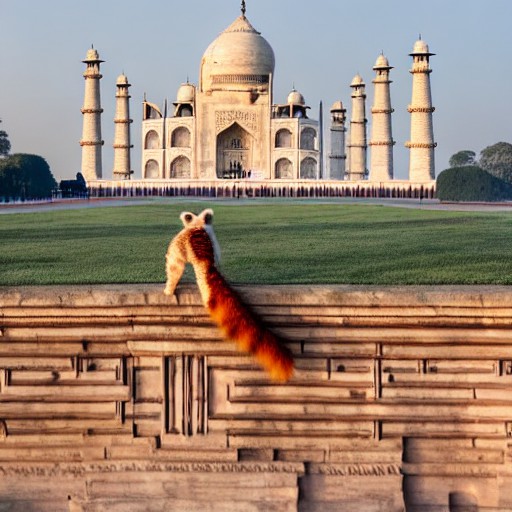}{
  A photo of red panda with white face markings and pointed ears in front of the Taj Mahal
}

\qualrownew{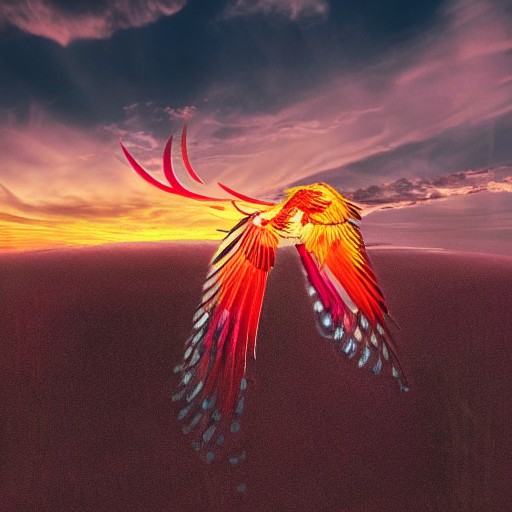}{
  A photo of phoenix with brilliant red and gold plumage in the sky during a vibrant sunset, with clouds painted orange and purple
}

\qualrownew{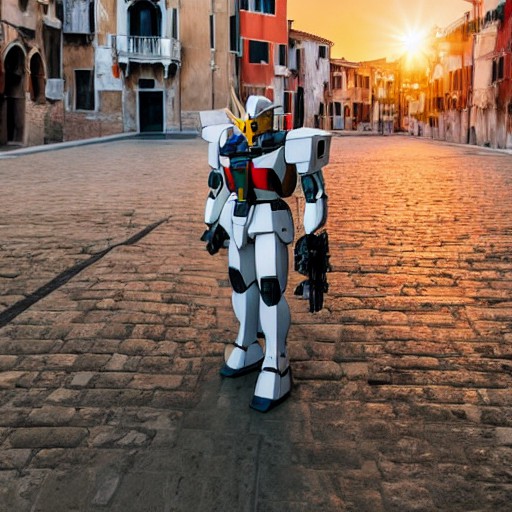}{
  A photo of mecha gundam with large shield and rifle on the streets of Venice, with the sun setting in the background
}


\bottomrule
\end{tabular}
\caption{\textbf{More qualitative comparisons (Part 2).} Each row pair shows images (top) and the corresponding prompt (bottom). The first two columns display the camera frustum (3D illustration) and a ground truth 3D rendering from a similar 3D object to the prompt. The remaining columns show results from different methods: ControlNet~\cite{zhang2023adding}, Stable-Virtual-Camera~\cite{zhou2025stable}, Compass Control~\cite{Parihar_2025_CVPR}, and our method. The object types in the prompts are not included in our training dataset.}
\label{fig:more_qualitative_grid_continued}
\end{figure*}

\begin{figure*}[t]
\small
\centering
\setlength{\tabcolsep}{1.5pt}
\renewcommand{\arraystretch}{1.12}
\begin{tabular}{ZYYYYY}
\toprule
\textbf{Camera Spec} &
\textbf{3D Render \hspace{5cm} (GT View)} &
\textbf{ControlNet} &
\textbf{SV-Camera} &
\textbf{Compass Control} &
\textbf{\;Ours\;} \\
\midrule

\qualrownew{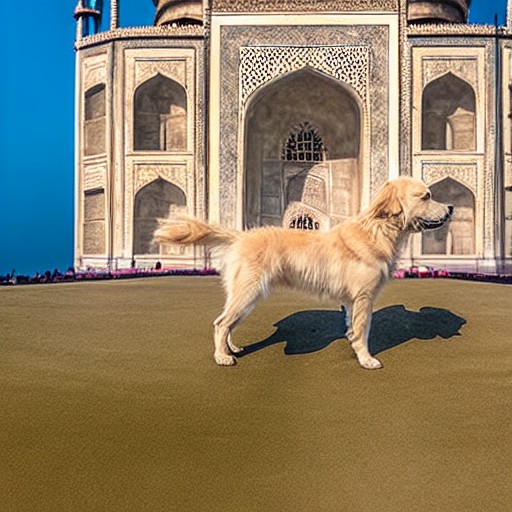}{
  A photo of golden retriever with fluffy fur in front of the Taj Mahal
}

\qualrownew{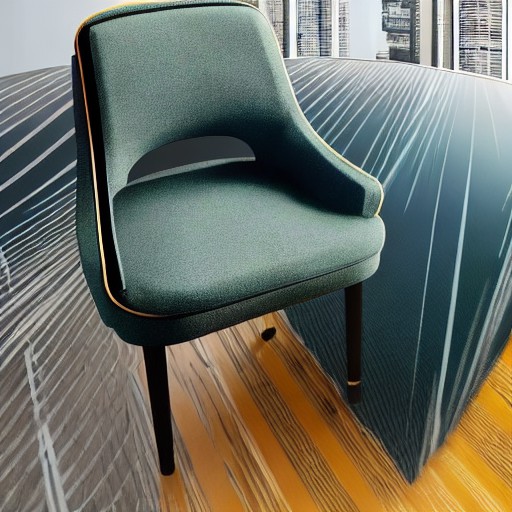}{
  A photo of ergonomic gaming chair with headrest in a modern living room setting with painted walls and glass windows
}

\qualrownew{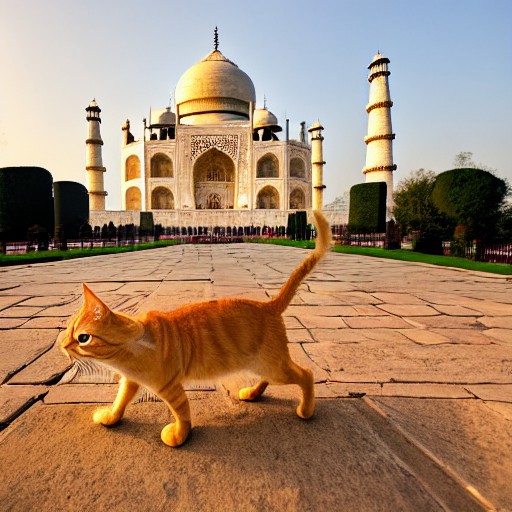}{
  A photo of orange tabby with green eyes and wearing a collar in front of the Taj Mahal
}

\qualrownew{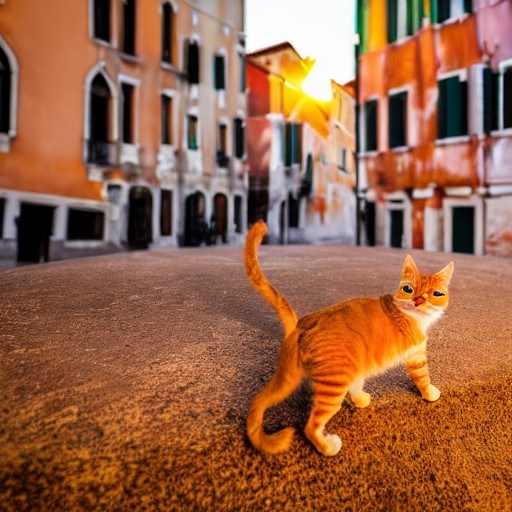}{
  A photo of orange tabby with green eyes and wearing a collar on the streets of Venice, with the sun setting in the background
}

\qualrownew{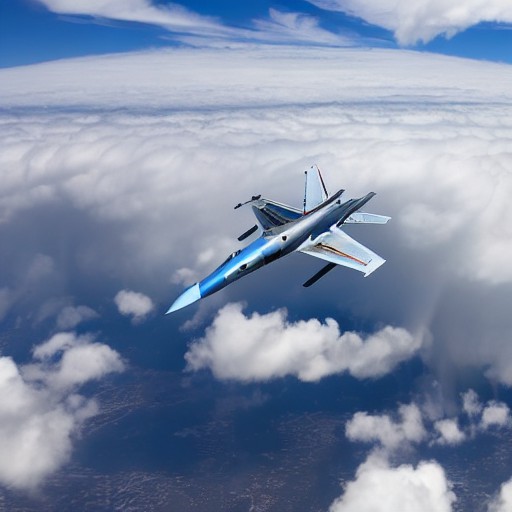}{
  A photo of fighter jet with afterburners and military markings flying high above a sea of fluffy white clouds
}

\bottomrule
\end{tabular}
\caption{\textbf{More qualitative comparisons (Part 3).} Each row pair shows images (top) and the corresponding prompt (bottom). The first two columns display the camera frustum (3D illustration) and a ground truth 3D rendering from a similar 3D object to the prompt. The remaining columns show results from different methods: ControlNet~\cite{zhang2023adding}, Stable-Virtual-Camera~\cite{zhou2025stable}, Compass Control~\cite{Parihar_2025_CVPR}, and our method. The object types in the prompts are included in our training dataset.}
\label{fig:more_qualitative_grid_3}
\end{figure*}

\end{document}